\documentclass[letterpaper]{article} 
\usepackage{aaai23}  
\usepackage{times}  
\usepackage{helvet}  
\usepackage{courier}  
\usepackage[hyphens]{url}  
\usepackage{graphicx} 
\usepackage{natbib}  
\usepackage{caption} 
\usepackage{algorithm}
\usepackage{algorithmic}

\usepackage{newfloat}
\usepackage{listings}
\DeclareCaptionStyle{ruled}{labelfont=normalfont,labelsep=colon,strut=off} 
\floatstyle{ruled}
\newfloat{listing}{tb}{lst}{}
\floatname{listing}{Listing}
\usepackage{amssymb}
\usepackage{amsthm}
\usepackage{mathtools}
\usepackage{physics}
\usepackage{makecell}
\usepackage{amsfonts,eucal}
\usepackage{bbold}
\usepackage{multirow}
\newcommand{\eat}[1]{}
\usepackage{wrapfig}
\usepackage[table]{xcolor}
\usepackage[utf8]{inputenc} 
\usepackage[T1]{fontenc}    
\usepackage{url}            
\usepackage{booktabs}       
\usepackage{amsfonts}       
\usepackage{nicefrac}       
\usepackage{microtype}      
\usepackage{xcolor}         
\usepackage{amsmath}
\usepackage{booktabs}
\usepackage{multirow}
\usepackage{graphicx}
\usepackage{bm}
\usepackage{enumitem}
\usepackage{framed}
\usepackage{amsmath}
\usepackage{multirow}
\newtheorem{theorem}{Theorem}[section]
\newtheorem{proposition}[theorem]{Proposition}

\newtheorem{definition}[theorem]{Definition}
\title{Handling Missing Data via Max-Entropy Regularized Graph Autoencoder}
\author{
   Ziqi Gao\textsuperscript{\rm 2},
      Yifan Niu\textsuperscript{\rm 1},
         Jiashun Cheng\textsuperscript{\rm 2},
        Jianheng Tang\textsuperscript{\rm 2},
              Tingyang Xu\textsuperscript{\rm 3},\\
        Peilin Zhao\textsuperscript{\rm 3},
        Lanqing Li\textsuperscript{\rm 3}\thanks{Corresponding author.},
        Fugee Tsung\textsuperscript{\rm 1,2},
        Jia Li\textsuperscript{\rm 1,2$\ast$}
}
\affiliations{
    \textsuperscript{\rm 1}The Hong Kong University of Science and Technology (Guangzhou)\\
    \textsuperscript{\rm 2}The Hong Kong University of Science and Technology\\
     \textsuperscript{\rm 3}AI Lab, Tencent\\
    zgaoat@connect.ust.hk, yniu669@connect.hkust-gz.edu.cn,\\ \{jchengak, jtangbf\}@connect.ust.hk, \{tingyangxu, masonzhao, lanqingli\}@tencent.com, \\ \{season, jialee\}@ust.hk
}

\usepackage{bibentry}

\begin{document}

\maketitle

\begin{abstract}
Graph neural networks~(GNNs) are popular weapons for modeling relational data. 
Existing GNNs are not specified for attribute-incomplete graphs, making missing attribute imputation a burning issue. Until recently, many works notice that
GNNs are coupled with spectral concentration, which means the spectrum obtained by GNNs concentrates on a local part in spectral domain, e.g., low-frequency due to oversmoothing issue. 
As a consequence, GNNs may be seriously flawed for reconstructing graph attributes as graph spectral concentration tends to cause a low imputation precision.
In this work, we present a regularized graph autoencoder for graph attribute imputation, named MEGAE, which aims at mitigating spectral concentration problem by maximizing the graph spectral entropy.
Notably, we first present the method for estimating graph spectral entropy without the eigen-decomposition of Laplacian matrix and provide the theoretical upper error bound.
A maximum entropy regularization then acts in the latent space, which directly increases the graph spectral entropy.
\eat{For a deeper understanding, we theoretically analyze and interpret the proposed autoencoder with the entropic regularization as a variational autoencoder (VAE) with a Uniform prior. 
Moreover, MEGAE avoids the problem of reparameterization and posterior collapse in VAE.}
Extensive experiments show that MEGAE outperforms all the other state-of-the-art imputation methods on a variety of benchmark datasets. 
\end{abstract}

\section{Introduction}
Graph attribute missing is ubiquitous due to messy collection and recording process~\cite{begin1,begin2}. 
In a biochemical scenario, for instance, \eat{data recorders may unintentionally omit some data entries~\cite{omit1,omit2,omit3}. More often,}data is missed due to the difficulties in measuring or calculating quantitative molecular properties (e.g., HOMO and LUMO orbital energy) at the atomic level~\cite{homo1,homo2}, attenuating the graph or node representations when we introduce a graph model for molecules.
For graph learning tasks, even a small fraction of missing attributes will potentially interfere with the performance, leading to biased inferences~\cite{domain1,domain3}.

\begin{figure}[t] 
    \centering
    \includegraphics[width=\linewidth]{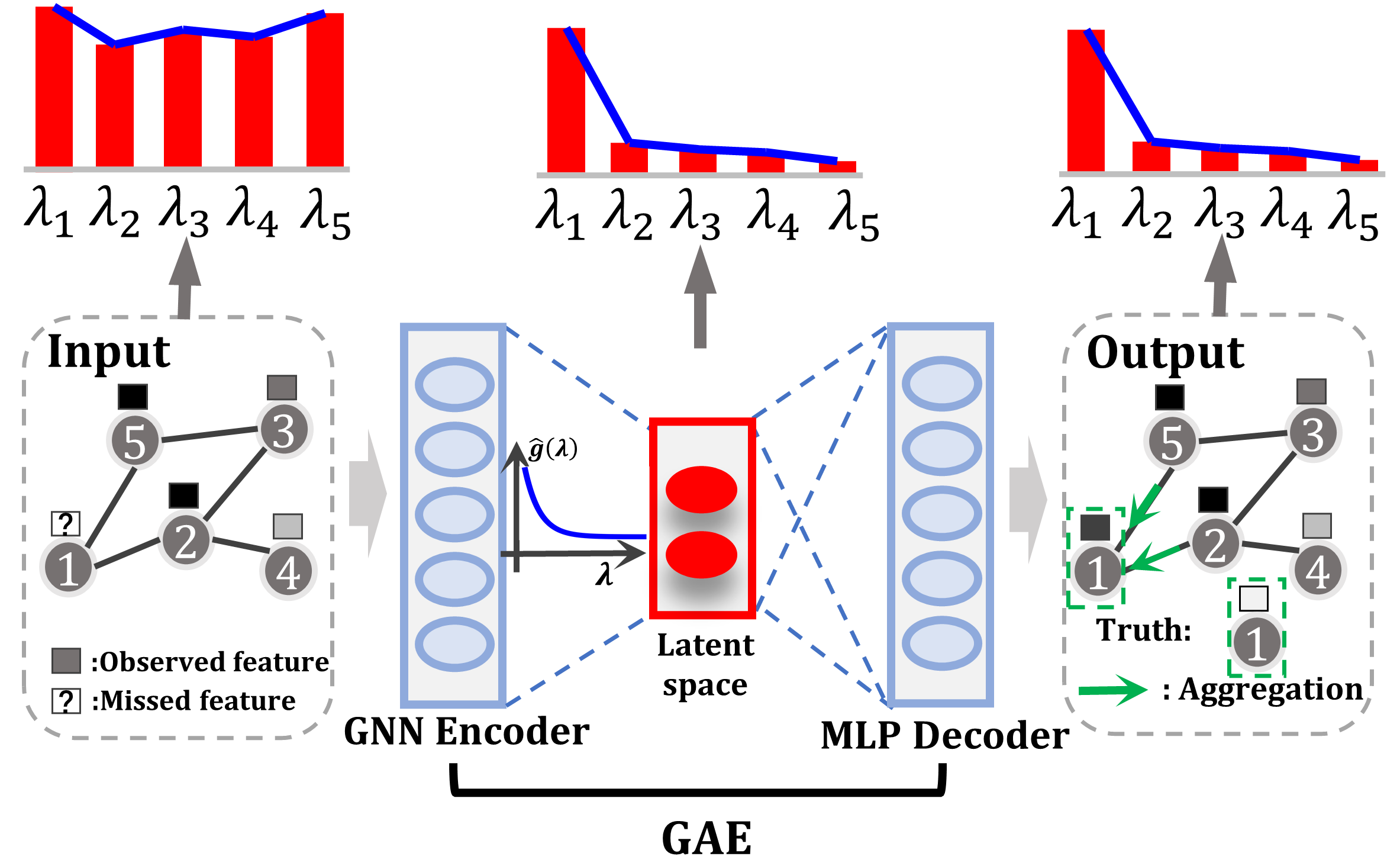}
    \caption{Spectral visualization of GAE. Red histograms show the graph spectra. After GAE encoder~(GCN), spectral concentration appears in the latent space. Upon MLP decoder, the spectral concentration is not alleviated.}
    \label{figure1}
\end{figure}

Graph Neural Networks~(GNNs) have demonstrated powerful ability on various graph-related tasks.
Most GNNs consist of two parts: feature propagation and transformation. The former is the key to the success of GNNs as it can yield suitable spectral responses for graph desires~\cite{g2cn}.
For example, graph convolutional networks (GCN)~\cite{gcn} applies a recursive aggregation mechanism for updating node representations, leading to the low-frequency concentration from a spectral view~\cite{sgcn,sgcn2,tang2022rethinking}.
The mechanism makes GCN effective for numerous real-world applications as low-frequency concentration promotes the convergence of representations of adjacent nodes~\cite{key1,key2}.
However, this spectral concentration diminishes partial eigenvalues, i.e., spectral values. As a consequence, it causes serious performance degradation on the standard graph attribute imputation models such as graph autoencoders~(GAEs)~\cite{li2020graph,gala,gdn}. In Figure~\ref{figure1}, we visualize the graph spectra of the input, latent and output based on GAE \cite{vgae}, which is the backbone of most GAEs. As the figure makes clear, existing GAEs leave sufficient room for improvement to combat the 
loss of spectral components. 
Delving into graph attribute reconstruction, a natural question arises: \textit{how can GAEs impute with alleviated spectral concentration?}
In this work, inspired by the maximum entropy principle~\cite{me_image1,me_image2,me_image3} that is capable to alleviate concentrated data distributions, we propose to \textit{maximize the graph spectral entropy} for mitigating spectral concentration while imputation.
We present a novel method for graph attribute imputation called Max-Entropy Graph AutoEncoder (MEGAE). Unlike existing imputation methods, our model trains a GAE with the regularization for maximizing the graph spectral entropy. 
However, when it comes to the computation of graph spectral entropy, it becomes challenging due to large time complexity of the eigen-decomposition for Laplacian matrix~\cite{eigen1,eigen2,eigen3}.
In this regard, we design the tight wavelet kernels~\cite{tight1,tight2,tight3} in MEGAE to encode features into the spectral domain involving several spectral bands.
We then theoretically prove that the proposed wavelet paradigm can well approximate the actual graph spectral entropy by a predictable upper error bound without eigen-decomposition. 
Upon entropy regularized latent representations, we use symmetric wavelet decoder to perform data deconvolution for reconstructing graph attributes. 


In the experiments, we proceed with an empirical
evaluation of MEGAE on single- and multi-graph datasets.
Firstly, we show that MEGAE outperforms 16 state-of-the-art methods on attribute imputation tasks, including commonly used classical methods and graph learning based imputation
methods. Additional downstream experiments of graph and node classifications demonstrate that imputed graphs by MEGAE can obtain the best accuracy performance when compared to state-of-the-art imputation models.
\paragraph{Contributions.}We summarize our contributions below:
\begin{itemize}
    \item We propose to maximize the graph spectral entropy to overcome the spectral concentration issue in GAEs for graph attribute reconstruction.
    \item We present Max-Entropy Graph AutoEncoder~(MEGAE) for encoding graphs into spectral domain and maximizing the spectral entropy in the latent space.
    \item We develop an efficient method for approximating the actual graph spectral entropy without eigen-decomposition, and importantly, provide the theoretical upper error bound.
\end{itemize}

\section{Related Work}
\paragraph{Matrix Completion.}Missing data is a widely researched topic. Matrix completion methods can be applied to impute graph attributes without using the graph structures.
Most proposed methods impute with a joint distribution on the incomplete data.
For instance, the joint modeling methods impute by drawing from the predictive distribution, including Bayesian strategies~\cite{bayesian}, matrix completion methods~\cite{mc1,mc2}, and Generative Adversarial Networks~\cite{gain,gan}. 
Another way of joint modeling involves iteratively imputing values of each variable using chained equations~\cite{mice} formulated with other variables~\cite{mice2, mice3,ot}. Discriminative models such as random forests~\cite{lp2}, distribution constraints using optimal transport~\cite{ot} and causally-aware imputation~\cite{miracle} tend to depend on strong assumptions, which may result in a  lack of flexibility to handle mixed-mode data. Most importantly, as for graph scenarios, matrix completion methods are generally limited without awareness of the underlying graph structures.  
\paragraph{Attribute Imputation with Graph Learning.}Recently, graph learning models have been used to tackle the imputation task. GC-MC~\cite{gcmc} and IGMC~\cite{gc2} construct helpful bipartite interaction graphs to impute with a given adjacency matrix as side information. Then a GAE is applied to predict the absent features. Gaussian mixture model is utilized for imputation stability under a high missing rate~\cite{gaussian}. GRAPE~\cite{grape} combines imputation and representation learning, which can well impute features of continuous variables but tend not to perform well on datasets containing nodes with all features missed~\cite{sia}. GDN~\cite{gdn} can impute from over-smoothed representations with a given graph structure. In this case, relationships between nodes are explicitly encoded for further imputation. Above methods show that graph learning is competent for feature reconstruction with structure dependencies. Notably, only GRAPE, GDN and our method are applicable to both discrete and continuous features. Despite both GDN and our method focus on mechanistic improvements for existing GAEs, 
we explore GAEs with a more general phenomenon of concentration in a spectral view and not just for recovering high-frequency details.

\section{Preliminary}\label{3}
\begin{figure*}[t] 
    \centering
    \includegraphics[width=\linewidth]{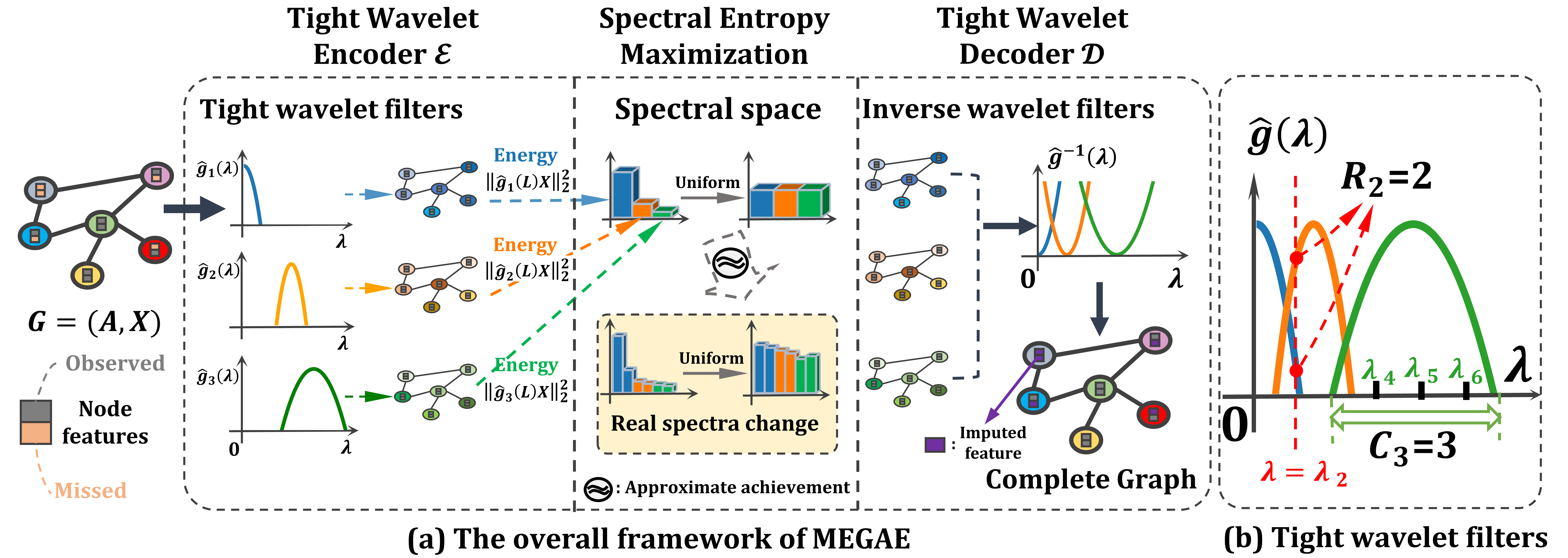}
    \caption{(a)~The proposed MEGAE for graph attribute imputation. With tight wavelet filters, the encoder filters the input graph into $\hat g_1(L)X, ..., \hat g_M(L)X$ ($M=3$ here), of which the energies~(i.e., blue, orange and green bars) are obtained with the L-2 norm function. We maximize wavelet entropy of $M$ energies, which are encouraged to be more uniform in distributions. This wavelet entropy maximization induces the same tendency to maximization for the real graph spectral entropy. We apply inverse wavelet filters to decode the data from the latent~(spectral) space for final reconstruction.~(b)~A diagram to explain the Coverage and Crossness of given tight wavelet filters. We provide related descriptions in Section~\ref{4.1}. }
    \label{figure2}
\end{figure*}
In this section, we provide the formulation of graph attribute imputation problem in Section~\ref{3.1}. Then we introduce the regularized object, named graph spectral entropy in Section~\ref{3.2}. In Section~\ref{3.3}, we describe wavelet entropy, a core concept for efficiently approximating the graph spectral entropy.

\subsection{Problem Formulation}\label{3.1}
For an undirected graph $G=(\bm{A},\bm{X})$, $\bm{A}\in \mathbb{R}^{N\times N}$ is the adjacency matrix and $\bm{X}\in \mathbb{R}^{N\times D}$ is a complete feature matrix where $\bm{X}_{ij}$ denotes the graph attribute of $i$-th node in the $j$-th feature dimension. 
In the problem of graph attribute imputation,  $\bm{R} \in \{ 0, 1 \}^{N \times D}$ is defined as the mask matrix whose element $\bm{R}_{ij}=1$ if $\bm{X}_{ij}$ is observed and $\bm{R}_{ij}=0$ otherwise.

The objective of this work is to predict the missing graph attributes $\bm{X}_{ij}$ at $\bm{R}_{ij}=0$. Formally, we aim to develop a mapping $f(\cdot)$ to generate the imputed data matrix $\tilde{\bm{X}} \in \mathbb{R}^{N \times D}$ defined as
\begin{equation}\label{eq1}
    \tilde{\bm{X}} = f(\bm{X}, \bm{R}, \bm{A}).
\end{equation}

\subsection{Graph Spectral Entropy}\label{3.2}
The \textit{graph Laplacian} is defined as $\bm{L}=\bm{I}-\tilde{\bm{A}}$, where $\tilde{\bm{A}}=\bm{D}^{-\frac{1}{2}}\bm{A}\bm{D}^{-\frac{1}{2}}$ denotes the normalized adjacency matrix and $\bm{D}=\text{diag}(\sum_k \bm{A}_{1k},...,\bm{A}_{Nk})$ denotes the diagnoal degree matrix. With eigen-decomposition, $\bm{L}$ can be decomposed into $\bm{L} = \bm{U}^{T} \bm{\Lambda} \bm{U}$, where $\bm{U}$ consists of the eigenvectors of $\bm{L}$. $\bm{\Lambda} = \text{diag}(\bm{\lambda})$ is the diagonal matrix whose diagonal elements $\bm{\lambda} = \{ \lambda_{1}, ..., \lambda_{N} \}$ are eigenvalues of $\bm{L}$. 
Consider a feature vector $\bm{x} = \{ x_{1}, ..., x_{N} \}^\intercal \in \mathbb{R}^{N}$ as a single feature dimension of the whole matrix $\bm{X}$, the Fourier transform $\bm{\hat{x}} = \bm{U}^{T} \bm{x} = \{ \hat{x}_{1}, ..., \hat{x}_{N} \}^\intercal$ would be the most proven method for representing spectrum, of which the entropy is defined as \textit{graph spectral entropy} in Definition~\ref{defi:gse}.
\begin{definition}[Graph Spectral Entropy] \label{defi:gse}
    Given the feature vector $\bm{x}$, its spectral energy at $\lambda_{i}$ is denoted as $\hat{x}_{i}^{2}$ and total energy is $E_{s} = \sum_{i=1}^{N} \hat{x}_{i}^{2} = \norm{\bm{U}^{T} \bm{x}}_{2}^{2}$. Then, the graph spectral entropy of $\bm{x}$ is defined as
    \[
    \xi_{s}(\bm{x}, \bm{L}) = -\sum_{i=1}^{N}\frac{\hat{x}_{i}^{2}}{E_{s}} \log \frac{\hat{x}_{i}^{2}}{E_{s}}.
    \]
\end{definition}
Intuitively, maximizing graph spectral entropy would ensure the spectral distribution to be relatively more uniform and keep information in any spectral pass-band, as opposed to spectral concentration in existing GAEs~\cite{gala,sgcn}.
However, one notable issue with the computation of graph spectral entropy is that eigen-decomposition is indispensable, which largely limits the scalability for computing on large-scale data matrix.  
\subsection{Wavelet Entropy}\label{3.3}
We introduce wavelet theory for approximating the graph spectral entropy with wavelet transform. Unlike Fourier transform that we formulate in section~\ref{3.2}, wavelet transform~\cite{gwnn, hammond} is able to \textit{cover ranges of spectra via different pass-band kernels} $\hat{g} = \{ \hat{g}_{1}, .., \hat{g}_{M} \}$. It shows the possibility of representing graph spectral entropy by wavelet entropy with well-calibrated kernels.

The wavelet transform employs a set of $M$ wavelets as bases, defined as $\mathcal{W} = \{ \mathcal{W}_{\hat{g}_1}, ..., \mathcal{W}_{\hat{g}_M} \}$. Mathematically, the wavelet transform of $\bm{x}$ \eat{by $\mathcal{W}_{\hat{g}_m}$} is defined as
\begin{equation} \label{gwt}
    \mathcal{W}_{\hat{g}_m}(\bm{x}) = \bm{U} \hat{g}_{m}(\bm{\Lambda}) \bm{U}^T \bm{x}, 
\end{equation}
where $\hat{g}_{m}(\cdot)$ is a wavelet kernel function defined in spectral domain. For computational efficiency, in many cases, kernel function $\hat{g}_{m}$ should be smooth and easily approximated by polynomials such that $\bm{U} \hat{g}_{m}(\bm{\Lambda}) \bm{U}^T = \hat{g}_{m}(\bm{ L})$.

\begin{definition}[Wavelet Entropy] \label{defi:twe}
    Given the feature vector $\bm{x}$ and a set of wavelet kernel functions $\hat{g} = \{ \hat{g}_{1}, ..., \hat{g}_{M} \}$, the total wavelet energy of $\bm{x}$ is $E_{w} = \sum_{m=1}^{M} \norm{ \mathcal{W}_{\hat{g}_m}(\bm{x})}_{2}^{2}$. Then, the wavelet entropy of $\bm{x}$ is defined as
    \[
    \xi_{w}(\bm{x}, \bm{L}, \hat{g}) = -\sum_{m=1}^{M}\frac{ \norm{\mathcal{W}_{\hat{g}_m}(\bm{x})}_{2}^{2}}{E_{w}} \log \frac{\norm{\mathcal{W}_{\hat{g}_m}(\bm{x})}_{2}^{2}}{E_{w}}.
    \]
\end{definition}

\section{Maximizing Graph Spectral Entropy}\label{4}
This section describes our main contributions. Following our motivation to efficiently maximize the graph spectral entropy while imputing missing attributes, we propose an alternative method for approximating the graph spectral entropy
\textit{without eigen-decomposition} and provide the upper bound for approximation error. On this basis, we propose our model called Max-Entropy Graph AutoEncoder (MEGAE).
\subsection{Graph Wavelet for Approximation} \label{4.1}
In this section, we present our main idea that graph spectral entropy can be well approximated by wavelet entropy. 
To achieve the goal, the crucial issue is how to construct a rough spectral space for approximation with the efficient graph convolutions. \eat{Given the unavailability of the implementation of the Fourier transform, graph wavelet, another excellent spectral kernel with higher sparsity and localization naturally becomes the most promising paradigm.}\eat{A wavelet paradigm is preferred for filtering the input graph with several pass-band kernels, which merges the spectra around each band center in a weighted manner.}
\begin{definition}[Tight Wavelet Frame on Graph] \label{defi:twf}
    A set of wavelet kernel functions $\hat{g} = \{ \hat{g}_{1}, ..., \hat{g}_{M} \}$ forms a tight frame on graph if
    \[
    G(\lambda_{i}) \coloneqq \sum_{m=1}^{M} [\hat{g}_{m}(\lambda_{i})]^{2} = 1, \forall \lambda_{i} \in \bm{\lambda}.
    \]
\end{definition}
Inspired by the fact that tight wavelet frame can ensure energy preservation~(Definition~\ref{defi:twf}) in graph signal processing~\cite{tight2}, we can smoothly deduce a key connection between wavelet energy~(formulated in Section~\ref{3.3}) and graph spectral energy~(formulated in Section~\ref{3.2}), which are both key components in the respective wavelet and spectral entropy. Proposition~\ref{propo1} describes this connection, indicating that the total wavelet energy is strictly equivalent to the total graph spectral energy.
The proof is trivial and illustrated in Appendix~\ref{prop1:proof}. 
\begin{proposition}[Energy Parseval's Identity] \label{propo1}
   If $\hat{g}$ is a tight wavelet frame, the total wavelet energy $E_{w}$ is equivalent to total graph spectral energy $E_{s}$. Thus, the Parseval's identity holds for the spectral domain, that is
  \[
        E_{w} = E_{s}.
  \]
\end{proposition}
The wavelet energy consists of $M$ energy components and each of them is derived from the filtered graph features by a wavelet kernel. Therefore, intuitively, a more distinguishable kernel function tends to provide accurate approximation of graph spectral entropy. To quantify the kernel distinguishability of tight wavelet, we define Coverage $C_{m}$ and Crossness $R_{i}$. Formally, $C_{m}$ is the number of spectrum covered by kernel function $\hat{g}_{m}$ denoted as
\begin{equation}
    C_{m} = \# \, \{i | \hat{g}_{m}(\lambda_i) \neq 0, i = 1, ..., N \},
\end{equation}
and $R_{i}$ is the number of activated kernel functions on $\lambda_i$ represented by
\begin{equation}
    R_{i} = \# \, \{m | \hat{g}_{m}(\lambda_i) \neq 0, m = 1, ..., M \}.
\end{equation}
A toy example is provided in Figure~\ref{figure2}(b) for easier understanding. Coverage $C_\textbf{3}$ represents the number of eigenvalues~(i.e., $\lambda_4,\lambda_5, \lambda_6$) lying within the domain of $\textbf{3}$rd wavelet kernel~(i.e., the green curve). The Crossness $R_\textbf{2}$ represents number of intersection~(red) points of all three wavelet kernels and line~$\lambda=\lambda_\textbf{2}$.   
\begin{proposition}[Approximation of graph spectral entropy]  \label{prop2}
     If $\hat{g}$ is a tight frame, the approximation error between graph spectral entropy $\xi_{s}(\bm{x}, \bm{L})$ and wavelet entropy $\xi_{w}(\bm{x}, \bm{L}, \hat{g})$ is bounded by
    \[
        |\xi_{s}(\bm{x}, \bm{L}) - \xi_{w}(\bm{x}, \bm{L}, \hat{g})| \leq e(\hat{g}, \bm{\lambda})
    \]
    where $e(\hat{g}, \bm{\lambda}) = \max(\{\log C_{m}\}_{m=1}^{M} \cup \{\log R_{i}\}_{i=1}^{N})$.
\end{proposition}
Please see Appendix~\ref{prop2:proof} for the proof. To better understand Proposition~\ref{prop2}, let $N=M$ and all kernels be disjoint, which means the number of wavelet kernels equals to the number of nodes in the graph. In this perfect situation, one kernel function only covers one eigenvalue~(i.e., Coverage $C_m=1$) and one eigenvalue is only covered by one kernel function~(i.e., Crossness $R_m=1$). Then, the approximation error is zero.

\eat{Based on Proposition~\ref{prop2}, we can obtain  well approximated graph spectral entropy by using a filter bank with kernels of sufficiently small bandwidth and overlapping support.}
For applying a tight wavelet frame with high distinguishability, we follow the mechanisms in~\cite{tight1} to design a set of wavelet kernel function $g=\{\hat{g}_1,...,\hat{g}_M \}$. The design details are in Appendix~\ref{twk:construction}.
\subsection{The Proposed MEGAE}
Based on the designed tight wavelet kernels, we propose MEGAE, a deterministic graph autoencoder framework for graph missing attribute imputation. As demonstrated in Figure~\ref{figure2}, MEGAE consists of two key modules: (1) wavelet based graph autoencoder and (2) maximum entropy regularization. 
\subsubsection{Encoder.}
With the definitions of an undirected graph provided in Section~\ref{3}, we employ $M$ tight wavelet kernels described in Section~\ref{4.1} in parallel to perform graph wavelet convolution. Specifically, the convolutional layer of $m$-th channel is expressed as
\begin{equation}
    \mathcal{E}_{1}(\bm{Z}^{(0)}_{m}, \bm{L}, \hat{g}_{m}) = \phi(\mathcal{W}_{\hat{g}_{m}}(\bm{Z}^{(0)}_{m}) \bm{W}_{m}^{(0)}),
\end{equation}
where $\bm{Z}^{(0)}_{m} = \bm{X} \odot \bm{R}$ and $\phi(\cdot)$ is the activation function such as leaky ReLU. To admit a fast algorithm for wavelet transform, Maclaurin series approximation~\cite{gdn} of order $K$ is applied such that $\mathcal{W}_{\hat{g}_m}(\bm{Z}^{(0)}_{m}) = \sum_{k=0}^{K} \alpha_{m, k} \bm{L}^{k} \bm{Z}^{(0)}_{m}$. We introduce one more layer of perceptron to enrich the representation in latent space, denoted as
\begin{equation}
    \mathcal{E}_{2}(\bm{Z}^{(1)}_{m}) = \phi(\bm{Z}^{(1)}_{m} \bm{W}_{m}^{(1)}).
\end{equation}
The \textit{Encoder} is thus formulated as
\begin{equation}
\begin{aligned}
    \bm{Z}^{(1)}_{m} & = \mathcal{E}_{1}(\bm{Z}^{(0)}_{m}, \bm{L}, \hat{g}_{m}), \\
    \bm{Z}^{(2)}_{m} & = \mathcal{E}_{2}(\bm{Z}^{(1)}_{m}).
\end{aligned}
\end{equation}
\subsubsection{Decoder.}
Given the observed information, our decoder aims to impute missing values from latent space. As the inverse of graph wavelet convolution, the graph wavelet deconvolutional layer of $m$-th channel is expressed as
\begin{equation}
    \mathcal{D}_{1}(\bm{Z}^{(2)}_{m}, \bm{L}, \hat{g}^{-1}_{m} ) = \phi(\mathcal{W}_{\hat{g}^{-1}_{m}}(\bm{Z}^{(2)}_{m}) \bm{W}_{m}^{(2)}),
\end{equation}
where $\mathcal{W}_{\hat{g}^{-1}_{m}}(\bm{Z}^{(2)}_{m}) = \sum_{k=1}^{K} \beta_{m, k} L^{k} \bm{Z}^{(2)}_{m}$ is provided by Maclaurin series. We then aggregate results generated by $M$ channels to construct the imputation layer as
\begin{equation}
    \mathcal{D}_{2}(\bm{Z}^{(3)}_{1}, ..., \bm{Z}^{(3)}_{M}) = \phi(\bm{Z}^{(3)}_{AGG} \bm{W}^{(3)}),
\end{equation}
where $\bm{Z}_{AGG}^{(3)} = \text{AGG}([\bm{Z}^{(3)}_{1}, ..., \bm{Z}^{(3)}_{M}])$ and $\text{AGG}(\cdot)$ denotes aggregation function such as concatenation.

The \textit{Decoder} is thus formulated as
\begin{equation}
\begin{aligned}
    \bm{Z}^{(3)}_{m} & = \mathcal{D}_{1}(\bm{Z}^{(2)}_{m}, \bm{L} , \hat{g}^{-1}_{m}), \\
    \tilde{\bm{X}} & = \mathcal{D}_{2}(\bm{Z}^{(3)}_{1}, ..., \bm{Z}^{(3)}_{M}).
\end{aligned}
\end{equation}

\subsubsection{Optimization.}
Our network is jointly optimized with respect to two objectives. We start from the imputation precision, which is the main focus in missing data problem. Intuitively, the reconstruction loss is defined as
\begin{equation}
    \mathcal{L}_{R} = \norm{(\tilde{\bm{X}} - \bm{X}) \odot (\bm{1}_{N \times D} - \bm{R})}_{2}.
\end{equation}
\begin{table*}[t]

\renewcommand\arraystretch{0.75}
\begin{center}

\begin{tabular}{@{}ccccccc|cc@{}}
\toprule
\multicolumn{1}{ p{2cm}<{\centering}|}{\multirow{2}{*}{\normalsize Methods}} & \multicolumn{1}{c|}{ENZYMES} & \multicolumn{1}{c|}{PRO\_full} & \multicolumn{1}{c|}{QM9} & \multicolumn{1}{c|}{Synthie} & \multicolumn{1}{c|}{FRANKE} & FIRST\_DB & \multicolumn{1}{c|}{ENZYMES} & PRO\_full \\ \cmidrule(l){2-9} 
\multicolumn{1}{c|}{} & \multicolumn{6}{c|}{RMSE with 0.1   missing features} & ACC. & ACC. \\ \midrule
MEAN & 0.0602 & 0.0653 & 0.2983 & 0.2063 & 0.3891 & 0.1500 & 62.06\% & 69.96\% \\
KNN & 0.0350 & 0.0329 & 0.3058 & 0.1718 & 0.2010 & 0.1296 & 63.53\% & 72.23\% \\
SVD & 0.0783 & 0.0465 & 0.2524 & 0.1697 & 0.2766 & 0.1685 & 61.60\% & 71.16\% \\
MICE & 0.0292 & 0.0210 & 0.1986 & 0.1899 & 0.1359 & 0.1036 & 64.46\% & 73.50\% \\
GAIN & 0.0300 & 0.0245 & 0.1973 & 0.1649 & 0.1103 & 0.0909 & 64.42\% & 73.21\% \\
OT & 0.0323 & 0.0206 & 0.2003 & 0.1865 & 0.1116 & 0.0892 & 64.13\% & 73.62\% \\
MIRACLE & 0.0288 & 0.0188 & 0.1846 & \underline{0.1632} & 0.1196 & 0.0889 & 65.03\% & 74.07\% \\
GraphVAE & 0.0357 & 0.0199 & 0.1579 & 0.1898 & 0.1099 & 0.1202 & 63.46\% & 73.86\% \\
MolGAN & 0.0326 & 0.0160 & \underline{0.1478} & 0.1864 & 0.1078 & 0.1379 & 64.16\% & 74.20\% \\
GRAPE & 0.0302 &  \underline{0.0147} & 0.1869 & 0.1798 & 0.1069 & 0.0986 & 64.48\% & \underline{74.53\%} \\
GDN & \underline{0.0267} & 0.0155 & 0.1598 & 0.1764 & \underline{0.1066} & \underline{0.0869} & \underline{65.57\%} & 74.50\% \\ \midrule
Inv\_GCN & 0.0256 & 0.0135 & 0.1773 & 0.1446 & 0.1089 & 0.1032 & 66.00\% & 74.98\% \\
Inv\_MLP & 0.0254 & 0.0129 & 0.1499 & 0.1520 & 0.1064 & 0.0872 & 66.06\% & 75.07\% \\
\textbf{MEGAE} & \textbf{0.0223} & \textbf{0.0099} & \textbf{0.1396} & \textbf{0.1203} & \textbf{0.0936} & \textbf{0.0789} & \textbf{66.27\%} & \textbf{75.97\%} \\ \midrule
\multirow{3}{*}{\begin{tabular}[c]{@{}c@{}}Performance\\ gain\end{tabular}} & \multirow{3}{*}{\begin{tabular}[c]{@{}c@{}}0.0044\\      |\\      0.0560\end{tabular}} & \multirow{3}{*}{\begin{tabular}[c]{@{}c@{}}0.0048\\      |\\      0.0554\end{tabular}} & \multirow{3}{*}{\begin{tabular}[c]{@{}c@{}}0.0082\\      |\\      0.1662\end{tabular}} & \multirow{3}{*}{\begin{tabular}[c]{@{}c@{}}0.0429\\      |\\      0.0860\end{tabular}} & \multirow{3}{*}{\begin{tabular}[c]{@{}c@{}}0.0128\\      |\\      0.2955\end{tabular}} & \multirow{3}{*}{\begin{tabular}[c]{@{}c@{}}0.0080\\      |\\      0.0896\end{tabular}} & \multirow{3}{*}{\begin{tabular}[c]{@{}c@{}}0.70\%\\      |\\      4.67\%\end{tabular}} & \multirow{3}{*}{\begin{tabular}[c]{@{}c@{}}1.44\%\\      |\\      6.01\%\end{tabular}} \\
 &  &  &  &  &  &  &  &  \\
 &  &  &  &  &  &  &  &  \\ \bottomrule
\end{tabular}%

 \end{center}
 \caption{RMSE results on 6 multi-graph datasets and graph classification accuracy on two datasets. \eat{The missing attributes are selected with a 10\% missing rate. }After running 5 trials, we report the mean results of which the best method is \textbf{bolded} and the second best is \underline{underlined}. Performance gains indicate the maximum~(lower) and minimum~(upper) gains of state-of-the-art~(MEGAE) compared to other baselines. Note that we abbreviate 'PROTEIN\_full', 'FRANKENSTEIN' and 'FIRSTMM\_DB' to 'PRO\_full', 'FRANKE' and 'FIRST\_DB', respectively. }
   \label{table1}
\end{table*}
Given our imputation motivation, MEGAE introduces spectral entropy regularization to aid data imputation. Based on the discussion in Section~\ref{4.1}, graph spectral entropy can be well substituted by wavelet entropy. Let $\bm{Z}_{m}^{(2)} (:, d)$ be the $d$-th dimension of $\bm{Z}_{m}^{(2)}$, the graph spectral entropy loss is thus defined as
\begin{equation}
    \mathcal{L}_{S} = - \frac{1}{D} \sum_{d=1}^{D} \sum_{m=1}^{M} P_{m}^{d} \log P_{m}^{d},
\end{equation}
where 
\begin{equation}
    P_{m}^{d} = \frac{\norm{\bm{Z}_{m}^{(2)} (:, d)}_{2}^{2}}{\sum_{m=1}^{M} \norm{\bm{Z}_{m}^{(2)} (:, d)}_{2}^{2}}.
\end{equation}
Finally, by combining the aforementioned losses, the overall objective function of MEGAE is formulated as
\begin{equation}\label{eq18}
    \mathcal{L} = \mathcal{L}_{R} - \gamma \mathcal{L}_{S},
\end{equation}
where $\gamma$ is a hyperparameter to balance the regularization of graph spectral entropy and reconstruction loss.

\section{Experiments}
In this section, we validate the performance of MEGAE using a variety of datasets. \eat{We denote MEGAE as our proposed imputation method with a given adjacency matrix and feature matrix. We conduct ablation study for analyzing MEGAE from perspectives of the wavelet design and overall loss for regularization.}
We evaluate the effectiveness of MEGAE on two categories of graph datasets:
\begin{itemize}
\item[$\bullet$] Type 1: \textbf{Imputation on multi-graph datasets.} We impute the missing graph attributes on multi-graph datasets, e.g., molecules, proteins. In addition, we report graph classification performance on graphs with imputed features.
\item[$\bullet$] Type 2: \textbf{Imputation on single-graph datasets.} We impute the missing values on single-graph datasets, e.g., social network. We report node classification performance on the graph with imputed features. 
\end{itemize}
\subsection{Imputation on Multi-graph Datasets.}\label{5.1}
\paragraph{Datasets.}\label{5.1.2}
We conduct experiments on 6 benchmark datasets~\cite{tud} from different domains: (1) bioinformatics, i.e., PROTEINS\_full~\cite{protein} and ENZYMES~\cite{enz}; (2) chemistry, i.e., QM9~\cite{qm91} and FIRSTMM\_DB~\cite{db1}; (3) computer vision, i.e., FRANKENSTEIN~\cite{frank}; (4) synthesis, i.e., Synthie~\cite{sy}. \eat{All datasets consist of continuous features and their sizes vary from $N = 10$ to $3848$. We preprocess the features with a MinMax scaler~\cite{scaler}.}The detailed statistics is provided in Table~\ref{table6} in Appendix~\ref{app:data}.
\paragraph{Baselines.}\label{5.1.3}
We compare the performance of MEGAE against baselines in three categories: (1) statistical imputation methods including MEAN, KNN~\cite{knn} and SVD~\cite{svd}; (2) deep learning-based imputation models including MICE~\cite{mice}, GAIN~\cite{gain}, OT~\cite{ot} and MIRACLE~\cite{miracle}; (3) graph learning-based models including GraphVAE~\cite{graphvae}, MolGAN~\cite{molgan}, GRAPE~\cite{grape} and GDN~\cite{gdn}. For further details, please refer to Appendix~\ref{app:baseline}.
\paragraph{Setup.}\label{5.1.1}
We use a 70-10-20 train-validation-test split and construct random missingness only on the test set. Each run has a different dataset split and the mask for feature missingness. \eat{Unless otherwise stated, missingness is applied by randomly removing features across all dimensions at a rate of 10\%.  We perform a hyperparameter sweep for $M$ and $\gamma$ as they change between $3$ and $20$ , $1e-2$ and $10$, respectively.} \eat{The values of hyperparameters are finally chosen by comparing the average imputation errors over all datasets. After imputation, we split the original training set into training and validation set by 7:1, and}After running for 5 trials, we report the Root Mean Squared Error (RMSE) results for imputation and mean accuracy on the test set for graph classification. For all baselines, we use a 2-layer GCN for downstream classification. For more setup details, please refer to Appendix~\ref{app:setup}.
\paragraph{Results.}\label{5.1.4}
\begin{table*}[t]
\centering
  \renewcommand\arraystretch{1.0}
\setlength\tabcolsep{2pt}
\begin{tabular}{@{}c|ccccc|ccccc@{}}
\toprule
\multirow{2}{*}{Methods} & \multicolumn{5}{c|}{PubMed (RMSE)} & \multicolumn{5}{c}{AmaPhoto (RMSE)} \\ \cmidrule(l){2-11} 
 & \multicolumn{1}{c|}{0.1 Miss} & \multicolumn{1}{c|}{0.3 Miss} & \multicolumn{1}{c|}{0.5 Miss} & \multicolumn{1}{c|}{0.7 Miss} & 0.99 Miss & \multicolumn{1}{c|}{0.1 Miss} & \multicolumn{1}{c|}{0.3 Miss} & \multicolumn{1}{c|}{0.5 Miss} & \multicolumn{1}{c|}{0.7 Miss} & 0.99 Miss \\ \midrule
sRMGCNN & \multicolumn{1}{c|}{0.0170} & \multicolumn{1}{c|}{{\underline{0.0186}}} & \multicolumn{1}{c|}{0.0258} & \multicolumn{1}{c|}{0.0309} & 0.0435 & \multicolumn{1}{c|}{0.3191} & \multicolumn{1}{c|}{0.3261} & \multicolumn{1}{c|}{{\underline{0.3282}}} & \multicolumn{1}{c|}{{\underline{0.3359}}} & 0.3660 \\
GC-MC & \multicolumn{1}{c|}{0.0220} & \multicolumn{1}{c|}{0.0233} & \multicolumn{1}{c|}{0.0253} & \multicolumn{1}{c|}{0.0270} & 0.0403 & \multicolumn{1}{c|}{0.3154} & \multicolumn{1}{c|}{0.3221} & \multicolumn{1}{c|}{0.3340} & \multicolumn{1}{c|}{0.3574} & 0.3913 \\
GRAPE & \multicolumn{1}{c|}{0.0187} & \multicolumn{1}{c|}{0.0198} & \multicolumn{1}{c|}{{\underline{0.0234}}} & \multicolumn{1}{c|}{0.0303} & 0.0368 & \multicolumn{1}{c|}{0.3207} & \multicolumn{1}{c|}{0.3295} & \multicolumn{1}{c|}{0.3357} & \multicolumn{1}{c|}{0.3489} & 0.3896 \\
VGAE & \multicolumn{1}{c|}{0.0196} & \multicolumn{1}{c|}{0.0229} & \multicolumn{1}{c|}{0.0298} & \multicolumn{1}{c|}{0.0354} & 0.0525 & \multicolumn{1}{c|}{0.3064} & \multicolumn{1}{c|}{0.3196} & \multicolumn{1}{c|}{0.3462} & \multicolumn{1}{c|}{0.3684} & 0.3991 \\
GDN & \multicolumn{1}{c|}{{\underline{0.0168}}} & \multicolumn{1}{c|}{0.0202} & \multicolumn{1}{c|}{0.0239} & \multicolumn{1}{c|}{{\underline{0.0254}}} & {\underline{0.0319}} & \multicolumn{1}{c|}{{\underline{0.3006}}} & \multicolumn{1}{c|}{{\underline{0.3168}}} & \multicolumn{1}{c|}{0.3308} & \multicolumn{1}{c|}{0.3372} & {\underline{0.3617}} \\
\textbf{MEGAE} & \multicolumn{1}{c|}{\textbf{0.0157}} & \multicolumn{1}{c|}{\textbf{0.0171}} & \multicolumn{1}{c|}{\textbf{0.0185}} & \multicolumn{1}{c|}{\textbf{0.0199}} & \textbf{0.0218} & \multicolumn{1}{c|}{\textbf{0.2966}} & \multicolumn{1}{c|}{\textbf{0.3121}} & \multicolumn{1}{c|}{\textbf{0.3204}} & \multicolumn{1}{c|}{\textbf{0.3243}} & \textbf{0.3305} \\ \midrule
\multirow{3}{*}{\begin{tabular}[c]{@{}c@{}}Performance\\       gain\end{tabular}} & \multicolumn{1}{c|}{\multirow{3}{*}{\begin{tabular}[c]{@{}c@{}}0.0011\\      |\\      0.0063\end{tabular}}} & \multicolumn{1}{c|}{\multirow{3}{*}{\begin{tabular}[c]{@{}c@{}}0.0015\\      |\\      0.0062\end{tabular}}} & \multicolumn{1}{c|}{\multirow{3}{*}{\begin{tabular}[c]{@{}c@{}}0.0049\\      |\\      0.0113\end{tabular}}} & \multicolumn{1}{c|}{\multirow{3}{*}{\begin{tabular}[c]{@{}c@{}}0.0055\\      |\\      0.0155\end{tabular}}} & \multirow{3}{*}{\begin{tabular}[c]{@{}c@{}}0.0101\\      |\\      0.0307\end{tabular}} & \multicolumn{1}{c|}{\multirow{3}{*}{\begin{tabular}[c]{@{}c@{}}0.0040\\      |\\      0.0241\end{tabular}}} & \multicolumn{1}{c|}{\multirow{3}{*}{\begin{tabular}[c]{@{}c@{}}0.0047\\      |\\      0.0174\end{tabular}}} & \multicolumn{1}{c|}{\multirow{3}{*}{\begin{tabular}[c]{@{}c@{}}0.0078\\      |\\      0.0258\end{tabular}}} & \multicolumn{1}{c|}{\multirow{3}{*}{\begin{tabular}[c]{@{}c@{}}0.0116\\      |\\      0.0441\end{tabular}}} & \multirow{3}{*}{\begin{tabular}[c]{@{}c@{}}0.0312\\      |\\      0.0686\end{tabular}} \\
 & \multicolumn{1}{c|}{} & \multicolumn{1}{c|}{} & \multicolumn{1}{c|}{} & \multicolumn{1}{c|}{} &  & \multicolumn{1}{c|}{} & \multicolumn{1}{c|}{} & \multicolumn{1}{c|}{} & \multicolumn{1}{c|}{} &  \\
 & \multicolumn{1}{c|}{} & \multicolumn{1}{c|}{} & \multicolumn{1}{c|}{} & \multicolumn{1}{c|}{} &  & \multicolumn{1}{c|}{} & \multicolumn{1}{c|}{} & \multicolumn{1}{c|}{} & \multicolumn{1}{c|}{} &  \\ \bottomrule
\end{tabular}%
\caption{Mean RMSE results of attribute imputation with different missing rates on PubMed and AmaPhoto.}
\label{table2}
\end{table*}
\begin{table*}[t]
\centering
  \renewcommand\arraystretch{1.0}
\setlength\tabcolsep{2pt}
\begin{tabular}{@{}c|ccccc|ccccc@{}}
\toprule
\multirow{2}{*}{Methods} & \multicolumn{5}{c|}{PubMed (Acc. = 77.11\% with full features)} & \multicolumn{5}{c}{AmaPhoto (Acc. = 92.03\% with full features)} \\ \cmidrule(l){2-11} 
 & \multicolumn{1}{c|}{0.1 Miss} & \multicolumn{1}{c|}{0.3 Miss} & \multicolumn{1}{c|}{0.5 Miss} & \multicolumn{1}{c|}{0.7 Miss} & 0.99 Miss & \multicolumn{1}{c|}{0.1 Miss} & \multicolumn{1}{c|}{0.3 Miss} & \multicolumn{1}{c|}{0.5 Miss} & \multicolumn{1}{c|}{0.7 Miss} & 0.99 Miss \\ \midrule
GCNMF & \multicolumn{1}{c|}{75.96\%} & \multicolumn{1}{c|}{74.42\%} & \multicolumn{1}{c|}{74.03\%} & \multicolumn{1}{c|}{70.09\%} & 41.46\% & \multicolumn{1}{c|}{91.64\%} & \multicolumn{1}{c|}{91.46\%} & \multicolumn{1}{c|}{91.05\%} & \multicolumn{1}{c|}{{\underline{90.16\%}}} & {\underline{73.03\%}} \\
PaGNN & \multicolumn{1}{c|}{{\underline{76.55\%}}} & \multicolumn{1}{c|}{{\underline{75.77\%}}} & \multicolumn{1}{c|}{{\underline{74.45\%}}} & \multicolumn{1}{c|}{72.15\%} & {\underline{43.63\%}} & \multicolumn{1}{c|}{91.81\%} & \multicolumn{1}{c|}{{\underline{91.70\%}}} & \multicolumn{1}{c|}{{91.02\%}} & \multicolumn{1}{c|}{90.12\%} & 71.51\% \\
sRMGCNN & \multicolumn{1}{c|}{75.96\%} & \multicolumn{1}{c|}{74.74\%} & \multicolumn{1}{c|}{72.79\%} & \multicolumn{1}{c|}{71.14\%} & 21.17\% & \multicolumn{1}{c|}{91.36\%} & \multicolumn{1}{c|}{91.19\%} & \multicolumn{1}{c|}{91.02\%} & \multicolumn{1}{c|}{90.12\%} & 66.56\% \\
GC-MC & \multicolumn{1}{c|}{74.44\%} & \multicolumn{1}{c|}{74.06\%} & \multicolumn{1}{c|}{72.87\%} & \multicolumn{1}{c|}{72.67\%} & 36.69\% & \multicolumn{1}{c|}{91.46\%} & \multicolumn{1}{c|}{91.26\%} & \multicolumn{1}{c|}{90.22\%} & \multicolumn{1}{c|}{89.82\%} & 60.13\% \\
GRAPE & \multicolumn{1}{c|}{75.53\%} & \multicolumn{1}{c|}{74.51\%} & \multicolumn{1}{c|}{73.65\%} & \multicolumn{1}{c|}{71.75\%} & 38.18\% & \multicolumn{1}{c|}{91.22\%} & \multicolumn{1}{c|}{91.05\%} & \multicolumn{1}{c|}{90.33\%} & \multicolumn{1}{c|}{89.63\%} & 68.28\% \\
VGAE & \multicolumn{1}{c|}{75.21\%} & \multicolumn{1}{c|}{73.69\%} & \multicolumn{1}{c|}{71.64\%} & \multicolumn{1}{c|}{69.96\%} & 16.66\% & \multicolumn{1}{c|}{91.53\%} & \multicolumn{1}{c|}{91.43\%} & \multicolumn{1}{c|}{89.23\%} & \multicolumn{1}{c|}{88.82\%} & 63.24\% \\
GDN & \multicolumn{1}{c|}{75.87\%} & \multicolumn{1}{c|}{74.22\%} & \multicolumn{1}{c|}{73.36\%} & \multicolumn{1}{c|}{{\underline{73.15\%}}} & 38.93\% & \multicolumn{1}{c|}{{\underline{91.83\%}}} & \multicolumn{1}{c|}{91.63\%} & \multicolumn{1}{c|}{\underline{91.10\%}} & \multicolumn{1}{c|}{90.01\%} & 66.62\% \\
\textbf{MEGAE} & \multicolumn{1}{c|}{\textbf{76.66\%}} & \multicolumn{1}{c|}{\textbf{76.30\%}} & \multicolumn{1}{c|}{\textbf{75.03\%}} & \multicolumn{1}{c|}{\textbf{74.47\%}} & \textbf{49.58\%} & \multicolumn{1}{c|}{\textbf{91.92\%}} & \multicolumn{1}{c|}{\textbf{91.85\%}} & \multicolumn{1}{c|}{\textbf{91.41\%}} & \multicolumn{1}{c|}{\textbf{90.58\%}} & \textbf{81.88\%} \\ \midrule
\multirow{3}{*}{\begin{tabular}[c]{@{}c@{}}Performance\\       gain\end{tabular}} & \multicolumn{1}{c|}{\multirow{3}{*}{\begin{tabular}[c]{@{}c@{}}0.11\%\\      |\\      1.45\%\end{tabular}}} & \multicolumn{1}{c|}{\multirow{3}{*}{\begin{tabular}[c]{@{}c@{}}0.53\%\\      |\\      2.61\%\end{tabular}}} & \multicolumn{1}{c|}{\multirow{3}{*}{\begin{tabular}[c]{@{}c@{}}0.58\%\\      |\\      3.39\%\end{tabular}}} & \multicolumn{1}{c|}{\multirow{3}{*}{\begin{tabular}[c]{@{}c@{}}1.32\%\\      |\\      4.51\%\end{tabular}}} & \multirow{3}{*}{\begin{tabular}[c]{@{}c@{}}5.95\%\\      |\\      32.92\%\end{tabular}} & \multicolumn{1}{c|}{\multirow{3}{*}{\begin{tabular}[c]{@{}c@{}}0.09\%\\      |\\      0.70\%\end{tabular}}} & \multicolumn{1}{c|}{\multirow{3}{*}{\begin{tabular}[c]{@{}c@{}}0.15\%\\      |\\      0.80\%\end{tabular}}} & \multicolumn{1}{c|}{\multirow{3}{*}{\begin{tabular}[c]{@{}c@{}}0.31\%\\      |\\      2.18\%\end{tabular}}} & \multicolumn{1}{c|}{\multirow{3}{*}{\begin{tabular}[c]{@{}c@{}}0.42\%\\      |\\      1.76\%\end{tabular}}} & \multirow{3}{*}{\begin{tabular}[c]{@{}c@{}}8.85\%\\      |\\      21.75\%\end{tabular}} \\
 & \multicolumn{1}{c|}{} & \multicolumn{1}{c|}{} & \multicolumn{1}{c|}{} & \multicolumn{1}{c|}{} &  & \multicolumn{1}{c|}{} & \multicolumn{1}{c|}{} & \multicolumn{1}{c|}{} & \multicolumn{1}{c|}{} &  \\
 & \multicolumn{1}{c|}{} & \multicolumn{1}{c|}{} & \multicolumn{1}{c|}{} & \multicolumn{1}{c|}{} &  & \multicolumn{1}{c|}{} & \multicolumn{1}{c|}{} & \multicolumn{1}{c|}{} & \multicolumn{1}{c|}{} &  \\ \bottomrule
\end{tabular}
\caption{Test accuarcy with different missing rates on PubMed and AmaPhoto node classification benchmarks.}
 \label{table3}
\end{table*}
In Table~\ref{table1} we show the experiment results for 10\% of missing features. Results indicate that our method has the lowest RMSE in the 5 real-world datasets and the synthetic dataset.
Moreover, almost all graph learning-based methods outperform non-graph learning ones, demonstrating the effectiveness of modeling the dependencies between graph nodes. In addition, we introduce two more baselines -- Inv\_MLP denotes our tight wavelet encoder with Multi-Layer Perceptron (MLP) decoder~\cite{graphvae} and Inv\_GCN denotes our encoder with GCN decoder~\cite{gcn}. \eat{We underline the second best methods as they perform the best other than `MEGAE-based' group.} MEGAE achieves 12.4\% lower RMSE than the second best baselines on average over all the datasets. 
For a more comprehensive and realistic consideration for missing mechanisms, we provide the evaluation~(see Figure~\ref{figure3} in Appendix~\ref{app:mar}) for evaluating MEGAE in two less common scenarios\eat{ for causing 30\% missingness}. We obtain similar gains across three missing machanisms, MEGAE obtains the lowest errors in all cases, which shows the strong generalization among the three missing mechanisms.

As for downstream classifications, we show the test accuracy of different models on imputed graph features with 10\% missing. We observe that smaller imputation errors are more likely to lead to higher test accuracy. This is relatively expected, as data with less distribution shift easily yields better representation. MEGAE outperforms all other baselines on accuracy conditioning on the effectiveness imputation results.
\subsection{Imputation on Single-graph Datasets.}\label{5.2}
\paragraph{Dataset.} We evaluate the effectiveness of MEGAE on four commonly used datasets including three citation network datasets---Cora, Citeseer, Pubmed~\cite{cora} and a co-purchase network dataset---AmaPhoto~\cite{amaphoto}. We show the dataset statistics in Table~\ref{table7} in Appendix~\ref{app:data}. 
\paragraph{Baselines.} We compare the imputation performance of MEGAE against competitive methods including sRMGCNN~\cite{mgcnn}, GC-MC~\cite{gcmc}, GRAPE~\cite{grape}, VGAE~\cite{vgae} and GDN~\cite{gdn}. For evaluating node classification performance of MEGAE, we additionally experiment with GCNMF~\cite{gcnmf} and PaGNN~\cite{pagnn} that directly implement graph learning without imputation. For further information, please refer to Appendix~\ref{app:baseline}.
\paragraph{Setup.} We closely follow \cite{gcn} to perform standard dataset split. \eat{As for each class, 20 nodes are randomly chosen for training. Moreover, we choose 500 labeled nodes for validation and evaluate the prediction accuracy on a test set of 1,000 labeled nodes. For AmaPhoto, we follow the work~\cite{gcnmf} to randomly choose 40 nodes for each class for training, 500 nodes for validation and the remaining nodes for testing.} For all datasets, each run has a different train-validation-test split and the mask for random missingness across all feature dimensions.\eat{ A 2-layer GCN~\cite{gcn} is used as a downstream GNNs for node classification.} After running for 5 trails, we report the results of mean RMSE for imputation and mean accuracy on the test set for node classification. For all baselines, we use a 2-layer GCN for downstream classification. For more setup details, please refer to Appendix~\ref{app:setup}.
\paragraph{Results.} Table~\ref{table2} shows the mean RMSE results for imputation under different missing rates on PubMed and AmaPhoto, we also provide the results on Cora and Citeseer in Table~\ref{table8} in Appendix. MEGAE consistently outperforms other baselines. Moreover, MEGAE tends to have greater performance gains when faced with a more severe missingness, which demonstrates MEGAE's robustness. It is noteworthy that under an extremely high missing rate~(0.99), MEGAE obtains margins of 31.66\% on PubMed, 8.63\% on AmaPhoto, 34.24\% on Cora and 19.54\% on Citeseer compared to the second best imputation methods.
In Table~\ref{table3}, we show the mean accuracy for node classification on PubMed and AmaPhoto with imputed attributes (results on Cora and Citeseer are in Table~\ref{table9} in Appendix). We observe that most methods work well with less than 0.3 missing rates. Moreover, MEGAE's robustness and effectiveness are better as its accuracy gains tend to be positively correlated with missing rates. With the consistent imputation improvements on all datasets, MEGAE has accuracy gains of 5.95\% on PubMed, 8.85\% on AmaPhoto, 7.90\% on Cora and 6.97\% on Citeseer under a 0.99 missing rate.
\subsection{Effectiveness of Wavelet and Regularization}
\paragraph{Graph Wavelet Autoencoders.}We study the role of tight wavelet frames with the configuration of regularization in MEGAE. Another two paradigms are deployed to get wavelet frames $\hat{g} = \{ \hat{g}_{1}, ..., \hat{g}_{M} \}$, including GWNN~\cite{gwnn} with heat kernels and SGWT~\cite{hammond} with highly-localized kernels.
\begin{table}[ht]	
\small
\centering
\begin{tabular}{cc|c|c}
\toprule
Dataset & GWNN & SGWT & Ours  \\
\midrule
ENZYMES         & +0.0042  & +0.0044  & \textbf{0.0223} \\
PROTEINS\_full  &  +0.0070 & +0.0052  & \textbf{0.0099} \\
QM9             &  +0.0102 & +0.0166  & \textbf{0.1396} \\
Synthie         &  +0.0203 & +0.0126  & \textbf{0.1203} \\
FRANKENSTEIN    &  +0.0103 & +0.0185  & \textbf{0.0936} \\
FIRST\_DB     & +0.0036  & +0.0073  & \textbf{0.0789} \\
\bottomrule
\end{tabular}
\caption{RMSE with three graph wavelet paradigms. We report the mean RMSE of MEGAE and show the increase w.r.t. RMSE of GWNN and SGWT compared to ours.}
\label{table4}
\end{table}

\begin{figure}[ht]
\centering
  \includegraphics[width=1.0\linewidth]{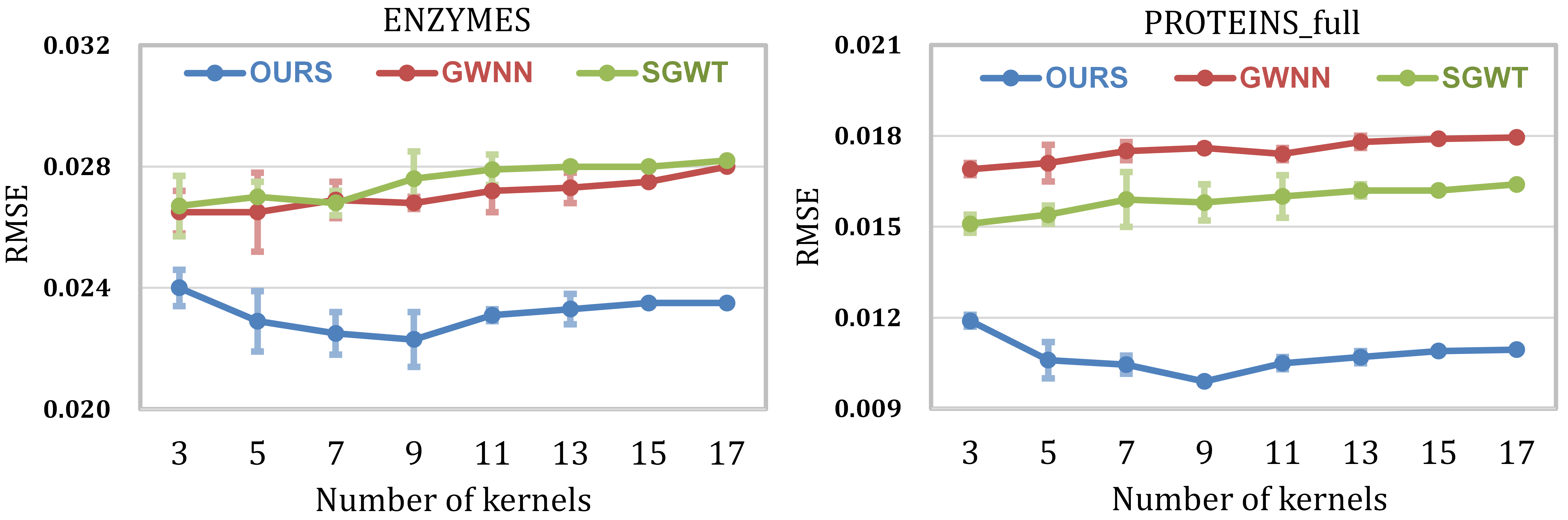}
\caption{RMSE with different numbers of wavelet kernels. We show how the kernel number influences the performance of GAEs with different wavelet paradigms.}
\label{figure4}
\end{figure}

Here we provide experiments on 6 multi-graph datasets under 0.1 missing rate (in Table \ref{table4}), reporting the increases w.r.t. RMSE using another two graph wavelet paradigms compared to ours, all three wavelet paradigms have 9 channels. 
Also, we execute experiments on 2 datasets (in Figure \ref{figure4}) assembling various channel numbers on three graph wavelet paradigms. 
Three paradigms all share the same 
computational complexity $O(K\times |\mathbb{E}|)$, where $K$ is the order of polynomials and $|\mathbb{E}|$ is the number of edges.
Table \ref{table4} verifies the effectiveness of the proposed tight wavelet kernels. Figure \ref{figure4} shows the influence of the number of wavelet kernels $M$. We observe that a larger $M$ value harms the performance of GAEs with GWNN and SGWT paradigms. As for our proposed MEGAE, the RMSE first drops and then increases along w.r.t. the number of wavelet kernels, which shows there is a tradeoff between entropy approximation error and model generalization.
\paragraph{Entropy Regularization.}We study the role of entropy regularization with the configuration of tight wavelet in MEGAE. We execute experiment on 6 datasets under 10\% missing rates. In Table \ref{table5} we provide the increase of RMSE by removing entropy regularization and the change in graph spectral entropy w.r.t. the original graph spectral entropy (denoted as $\Delta$). We observe that GAE without regularization will seriously lead to a decrease in spectral entropy~(i.e., aforementioned spectral concentration) of the imputed graphs, and more importantly, $\mathcal{L}_{S}$ improves the performance of the proposed MEGAE with a substantially 
maintenance of the real entropy. 
\begin{table}[ht]


\centering
\begin{tabular}{ccc|cc}
\toprule
Dataset & $\mathcal{L}_{R}$ & $\Delta_1$ & $\mathcal{L}_{R}+\mathcal{L}_{S}$ &  $\Delta_2$ \\
\midrule
ENZYMES       & +0.0069 & -29\% & \textbf{0.0223} & +3\% \\
PROTEINS\_full & +0.0061 &  -26\% & \textbf{0.0099} & +2\% \\
QM9           & +0.0188 &  -18\%  & \textbf{0.1396}  &  -1\%  \\
Synthie       & +0.0316 &  -23\%  & \textbf{0.1203} &  +3\% \\
FRANKENSTEIN  &   +0.0229 &  -33\% & \textbf{0.0936}  &  -2\% \\
FIRSTMM\_DB    &  +0.0086 & -21\% & \textbf{0.0789}  &  -2\% \\
\bottomrule
\end{tabular}
\caption{RMSE with/without entropy regularization. $\Delta_1$ and $\Delta_2$ denote the entropy changes with/without the regularization w.r.t. the original graph spectral entropy.}
\label{table5}
\end{table}
\paragraph{Discussion.}
The effectiveness of our proposed tight wavelet and entropy regularization has been shown in Table~\ref{table4} and Table~\ref{table5}, from which we can easily find that both have obvious improvements for decreasing imputation errors. Importantly, we believe the entropy regularization contributes the most to the performance of MEGAE.
By analogizing MEGAE to the variational autoencoders, we provide intuitions on why MEGAE works (see Appendix~\ref{app:vae}).

\section{Conclusion.}
In this paper we show that graph spectral entropy is a good optimizable proxy for graph attribute imputation. 
Motivated by eliminating the spectral concentration of GAEs and retaining all spectral components, we propose MEGAE with a graph spectral entropy regularization for imputation. We develop a theoretical guarantee that our proposed MEGAE could maximize the target entropy without eigen-decomposition.
Experimental results of various datasets on imputation and downstream classification tasks show the effectiveness of MEGAE for handling missing attributes on graphs.

\section*{Acknowledgements}
The research of Li was supported by NSFC Grant No. 62206067, Tencent AI Lab Rhino-Bird Focused Research Program RBFR2022008 and Guangzhou-HKUST(GZ) Joint Funding Scheme. The research of Tsung was supported by the Hong Kong RGC General Research Funds 16216119 and Foshan HKUST Projects FSUST20-FYTRI03B.

\bibliography{aaai23}

\appendix
\clearpage
\appendix\label{app}
\section*{Technical Appendix}
In the technical appendix, we provide further proof for the propositions as well as details for the experiments.
This appendix is outlined as follows:
\begin{itemize}
    \item Section~\ref{prop1:proof} provides the proof for Proposition~\ref{propo1}.
    \item Section~\ref{prop2:proof} provides the proof for Proposition~\ref{prop2}.
    \item Section~\ref{twk:construction} details the approach to construct tight wavelet kernels applied in this paper.
    \item Section~\ref{app:vae} provides a novel intuition from a VAE perspective for explaining why MEGAE works. Derivations are also provided to corroborate the intuition.
    \item Section~\ref{app:exp} provides the experimental details including dataset summary, setup, baseline implementations and computational hardware.
    \item Section~\ref{app:mar} provides additional experiments on multi-graph datasets testing for imputation performance. We consider MCAR, MAR and MNAR missingness patterns.
    \item Section~\ref{app:cora} provides additional experiments for imputation and downstream classification results on Cora and Citeseer.
\end{itemize}
\section{Proof of Proposition~\ref{propo1}}  \label{prop1:proof}
\begin{proof}
    Considering the fact that $\hat{g}$ is a tight frame, the wavelet energy can be transformed into
    \begin{equation*}
    \begin{aligned}
        E_{w} & = \sum_{m=1}^{M} \norm{ \mathcal{W}_{\hat{g}_m}(\bm{x})}_{2}^{2} \\
        & = \sum_{m=1}^{M} \sum_{i=1}^{N}  \hat{g}_{m}^{2}(\lambda_{i}) \hat{x}_{i}^{2} \\
        & = \sum_{i=1}^{N} (\sum_{m=1}^{M} \hat{g}_{m}^{2}(\lambda_{i})) \hat{x}_{i}^{2} \\
        & = \sum_{i=1}^{N} \hat{x}_{i}^{2} \\
        & = E_{s}.
    \end{aligned}
    \end{equation*}
    
    Therefore, Parseval identity holds for the spectral energy and wavelet energy.
\end{proof}

\section{Proof of Proposition~\ref{prop2}} \label{prop2:proof}

\begin{proof}
    For simplicity of notations, the graph spectral entropy $\xi_{s}(\bm{x}, \bm{L})$ and wavelet entropy $\xi_{w}(\bm{x}, \bm{L}, \hat{g})$ can be denoted as
    \begin{equation*}
    \begin{aligned}
        \xi_{s}(\bm{x}, \bm{L}) & = - \sum_{i=1}^{N} p_{i} \log p_{i}, \\
        \xi_{w}(\bm{x}, \bm{L}, \hat{g}) & = 
        - \sum_{m=1}^{M} q_{m} \log q_{m},
    \end{aligned}
    \end{equation*}
    where $p_{i} = \frac{\hat{x}_{i}^{2}}{E_{s}}$ and $q_{m} = \frac{\norm{\mathcal{W}_{\hat{g}_m}(\bm{x})}_{2}^{2}}{E_{w}}$. With Proposition~\ref{propo1}, $E_w$ can be replaced by $E_s$, thus $q_{m}$ can be further represented as
    \begin{equation*}
    \begin{aligned}
        q_{m} & = \frac{\sum_{i=1}^{N} \hat{g}_{m}^{2}(\lambda_i) \hat{x}_{i}^{2}}{E_{s}} \\
        & = \sum_{i=1}^{N} \hat{g}_{m}^{2}(\lambda_i) p_{i}.
    \end{aligned}    
    \end{equation*}
    
    Note that $q_{m}$ is a weighted combination of $p_{i}$ with $m$-th wavelet kernel, we further decompose $q_{m}$ and define the auxiliary energy $k_{i}^{m} = \hat{g}_{m}^{2}(\lambda_i) p_{i}$ with auxiliary entropy
    \begin{equation*}
        - \sum_{i=1}^{N} \sum_{m=1}^{M} k_{i}^{m} \log k_{i}^{m} = - \sum_{i=1}^{N} \sum_{m=1}^{M} \hat{g}_{m}^{2}(\lambda_i) p_{i} \log \hat{g}_{m}^{2}(\lambda_i) p_{i}.
    \end{equation*}
    
    Based on the definition of tight frame, it can be shown $\hat{g}_{m}^{2}(\lambda_i) \leq 1$. Therefore, graph spectral entropy and wavelet entropy are less than auxiliary entropy with
    \begin{equation*}
    \begin{aligned}
        - \sum_{i=1}^{N} p_{i} \log p_{i} & = - \sum_{i=1}^{N} \sum_{m=1}^{M} \hat{g}_{m}^{2}(\lambda_i) p_{i} \log p_{i} \\
        & \leq - \sum_{i=1}^{N} \sum_{m=1}^{M} k_{i}^{m} \log k_{i}^{m}, \\
        - \sum_{m=1}^{M} q_{m} \log q_{m} & = - \sum_{i=1}^{N} \sum_{m=1}^{M} \hat{g}_{m}^{2}(\lambda_i) p_{i} \log \sum_{j=1}^{N} \hat{g}_{m}^{2}(\lambda_j) p_{j} \\
        & \leq - \sum_{i=1}^{N} \sum_{m=1}^{M} k_{i}^{m} \log k_{i}^{m}.
    \end{aligned}
    \end{equation*}
    
    \textbf{Condition 1:} $ - \sum_{i=1}^{N} p_{i} \log p_{i} < - \sum_{m=1}^{M} q_{m} \log q_{m}$.
    
    By the inequalities of auxiliary entropy, it's trivial to derive
    \begin{equation*}
    \begin{aligned}
        & | \sum_{i=1}^{N} p_{i} \log p_{i} - \sum_{m=1}^{M} q_{m} \log q_{m} | \\ 
        & \leq | \sum_{i=1}^{N} p_{i} \log p_{i} - \sum_{i=1}^{N} \sum_{m=1}^{M} k_{i}^{m} \log k_{i}^{m} | \\
        & = \sum_{i=1}^{N} p_{i} \log p_{i} - \sum_{i=1}^{N} \sum_{m=1}^{M} k_{i}^{m} \log k_{i}^{m} \\
        & = \sum_{i=1}^{N} p_{i} \sum_{m=1}^{M} \hat{g}_{m}^{2}(\lambda_i) \log \frac{1}{\hat{g}_{m}^{2}(\lambda_i)}.
    \end{aligned}
    \end{equation*}
    
    Denote $\tilde{R}_{i} = \{m | \hat{g}_{m}(\lambda_i) \neq 0, m = 1, ..., M \}$ as the index set of activated kernel functions on $\lambda_{i}$ and $R_{\max} = \max(\{ R_{i}  \}_{i=1}^{N})$. Note that $\phi(x) = x \log \frac{1}{x}$ is a concave function, we proceed the derivation by applying Jensen inequality to the last step as
    \begin{equation*}
    \begin{aligned}
        & \sum_{i=1}^{N} p_{i} \sum_{m=1}^{M} \hat{g}_{m}^{2}(\lambda_i) \log \frac{1}{\hat{g}_{m}^{2}(\lambda_i)} \\
        & = \sum_{i=1}^{N} p_{i} \sum_{m \in \tilde{R}_{i}} \phi(\hat{g}_{m}^{2}(\lambda_i)) \\
        & \leq \sum_{i=1}^{N} p_{i} \phi(\frac{\sum_{m \in \tilde{R}_{i}} \hat{g}_{m}^{2}(\lambda_i)}{R_{i}}) R_{i} \\ 
        & \leq \sum_{i=1}^{N} p_{i} \log R_{\max} \\
        & = \log R_{\max}.
    \end{aligned}
    \end{equation*}
    
    Combining the two parts, we have shown that
    \begin{equation*}
        | \sum_{i=1}^{N} p_{i} \log p_{i} - \sum_{m=1}^{M} q_{m} \log q_{m} | \leq \log R_{\max}.
    \end{equation*}
    
    \textbf{Condition 2:} $ - \sum_{i=1}^{N} p_{i} \log p_{i} > - \sum_{m=1}^{M} q_{m} \log q_{m}$.
    
    Similarly, denote $\tilde{C}_{m} = \{i | \hat{g}_{m}(\lambda_i) \neq 0, i = 1, ..., N \}$ as the index set of spectrum covered by kernel function $\hat{g}_{m}$ and $C_{\max} = \max(\{ C_{m} \}_{m=1}^{M})$. By definition, $\sum_{m=1}^{M} q_{m} \log q_{m}$ can be represented by convex function $\psi(x) = x \log x$. By Jensen inequality, graph wavelets entropy is interpreted as
    \begin{equation*}
    \begin{aligned}
        & \sum_{m=1}^{M} q_{m} \log q_{m} \\
        & = \sum_{m=1}^{M} \sum_{i=1}^{N} \hat{g}_{m}^{2}(\lambda_i) p_{i} \log \sum_{j=1}^{N} \hat{g}_{m}^{2}(\lambda_j) p_{j} \\
        & = \sum_{m=1}^{M} \psi(\frac{\sum_{i \in \tilde{C}_{m}} \hat{g}_{m}^{2}(\lambda_i) p_{i}}{C_{m}} C_{m}) \\
        & \leq \sum_{m=1}^{M} \sum_{i=1}^{N} \hat{g}_{m}^{2}(\lambda_i) p_{i} \log  \hat{g}_{m}^{2}(\lambda_i) p_{i} C_{m}
    \end{aligned}
    \end{equation*}
    
    According to the inequalities of auxiliary entropy, we can get
    \begin{equation*}
    \begin{aligned}
        & | \sum_{i=1}^{N} p_{i} \log p_{i} - \sum_{m=1}^{M} q_{m} \log q_{m} | \\ 
        & \leq | \sum_{m=1}^{M} q_{m} \log q_{m} - \sum_{i=1}^{N} \sum_{m=1}^{M} k_{i}^{m} \log k_{i}^{m} | \\
        & = \sum_{m=1}^{M} q_{m} \log q_{m} - \sum_{i=1}^{N} \sum_{m=1}^{M} k_{i}^{m} \log k_{i}^{m} \\
        & \leq \sum_{m=1}^{M} \sum_{i=1}^{N} \hat{g}_{m}^{2}(\lambda_i) p_{i} \log \frac{\hat{g}_{m}^{2}(\lambda_i) p_{i} C_{m}}{\hat{g}_{m}^{2}(\lambda_i) p_{i}} \\
        & \leq \sum_{m=1}^{M} \sum_{i=1}^{N} \hat{g}_{m}^{2}(\lambda_i) p_{i} \log C_{\max} \\
        & = \log C_{\max}.
    \end{aligned}
    \end{equation*}
    
    In conclusion, we prove that
    \begin{equation*}
        |\xi_{s}(\bm{x}, \bm{L}) - \xi_{w}(\bm{x}, \bm{L}, \hat{g})| \leq e(\hat{g}, \bm{\lambda})
    \end{equation*}
    where $e(\hat{g}, \bm{\lambda}) = \max(\log C_{\max}, \log R_{\max})$.
\end{proof}
\section{Tight Wavelet Kernel Construction} \label{twk:construction}
Here, we follow the approach proposed in~\cite{tight1}. As stated in Section~\ref{4.1}, let $T$ and $Q$ be any integers satisfying $2 < T \leq M$ and $Q < \frac{T}{2}$, the mother wavelet is defined as
\begin{equation*}
    \hat{g}^{U}(\lambda) = \sum_{q=0}^{Q} a_{q} 
    \begin{bmatrix} 
        \cos(2\pi q(\frac{M+1-T}{T\gamma} + \frac{1}{2})) \\ 
        \cdot \mathbb{1}_{\{-\frac{R\gamma}{M+1-R} \leq \lambda<0\}}
    \end{bmatrix},
\end{equation*}
where $\gamma$ is the upper bound on spectrum and coefficients $\{ a_{q} \}_{q=1}^{Q}$ is a real sequence satisfying $\sum_{q=0}^{Q} (-1)^{q} a_{q} = 0$.

By introducing graph translation, we obtain the preliminary tight wavelet kernel functions
\begin{equation*}
    \hat{g}^{U}_{m}(\lambda) = \hat{g}^{U}(\lambda - m\frac{\gamma}{M+1-T}),
\end{equation*}
satisfying 
\begin{equation*}
    G^{U}(\lambda) = \sum_{m=1}^{M} [\hat{g}^{U}_{m}(\lambda)]^2 = T a_{0}^{2} + \frac{T}{2} \sum_{q=0}^{Q} a_{q}^{2}, \forall \lambda \in \bm{\lambda}.
\end{equation*}

We consider wrapping function $\omega(\lambda)=\log(\lambda)$ instead of linear scale transformation. Eventually, the tight wavelet kernel functions are defined as
\begin{equation*}
\begin{aligned}
    \hat{g}_{m}(\lambda) & \coloneqq \hat{g}^{U}_{m}(\omega(\lambda)), \, m = 2, ..., M, \\
    \hat{g}_{1}(\lambda) & \coloneqq \sqrt{G^{U}(\lambda) - \sum_{m=2}^{M} [\hat{g}_{m}(\lambda)]^{2}}.
\end{aligned}
\end{equation*}

\section{Relationship to Variational Autoencoder}\label{app:vae}
\subsection{Resemblances with VAE}
We derived our formulation directly from the spectral entropy maximization problem. However, once the assumption of prior distribution $p\left(\bm{Z}\right)$ is made, the optimization objective resembles those in stochastic variational inference and learning algorithm, and can be projected more specifically into variational autoencoders (VAE)~\cite{vae}. VAE aims to learn an encoder and a decoder to map the input space $\bm{X}\in \bm{\mathcal{X}}$ to the stochastic embedding $\bm{Z} \in \bm{\mathcal{Z}}$. We define the variational posterior $q_\phi(\bm{Z}|\bm{X})$ and the decoder by a generative distribution $p_\theta(\bm{X}|\bm{Z})$, where $\phi$ and $\theta$ are learned parameters.  
The optimization objective is the variational lower bound $\mathcal{L}$ w.r.t. the variational parameters $\phi,\theta$:
\begin{equation}
\begin{aligned}
\mathcal{L}(\theta,\phi;\bm{X})=\mathbb{E}_{q_\phi(\bm{Z}|\bm{X})}[&-log p_\theta(\bm{X}|\bm{Z})]\\
&+\text{KL}[ q_\phi(\bm{Z}|\bm{X})||p(\bm{Z})],
\end{aligned}
\end{equation}
Consider a prior distribution of isotropic Uniform, i.e. $\bm{Z} \sim U[\bm{a}, \bm{b}]$,  imposed on the latent code representation, the optimization objective can be reformulated as:

\begin{equation}\label{eqvae}
\begin{aligned}
\mathcal{L}(\theta,\phi;\bm{X})=&\mathbb{E}_{q_\phi(\bm{Z}|\bm{X})}[-log p_\theta(\bm{X}|\bm{Z})]-\text{H}[ q_\phi(\bm{Z}|\bm{X})]\\
&+\text{H}[ q_\phi(\bm{Z}|\bm{X}), p(\bm{Z})],
\end{aligned}
\end{equation}

where $\text{H}(q)$ is the entropy of $q$,  $\text{H}(q,p)$ is the cross-entropy of the distribution $p$ relative to a distribution $q$ over a given set. Given the prior distribution $p(\bm{Z})$, we can explicitly derive the cross entropy $\text{H}[ q_\phi(\bm{Z}|\bm{X}), p(\bm{Z})]$ as:

\begin{equation}\label{eq17}
\begin{aligned}
    \text{H}[ q_\phi(\bm{Z}|\bm{X}), p(\bm{Z})] &= \int_a^b q_\phi(\bm{Z}|\bm{X})\text{log}p(\bm{Z})d\bm{Z}\\
    &=C\int_a^b q_\phi(\bm{Z}|\bm{X})d\bm{Z}=C,
    \end{aligned}
\end{equation}

where $C$ is a finite constant. Therefore, we omit this constant term in Eq.~\ref{eq17}. Eventually, the optimization objective of VAE is defined as:

\begin{equation}\label{eqvae1}
\mathcal{L}(\theta,\phi;\bm{X})=\mathbb{E}_{q_\phi(\bm{Z}|\bm{X})}[-log p_\theta(\bm{X}|\bm{Z})]-\text{H}[ q_\phi(\bm{Z}|\bm{X})].
\end{equation}

This variant objective function of VAE is equivalent to our data imputation problem based on graph spectral entropy maximization, as defined in Eq.~\ref{eq18}. With a prior distribution of isotropic Uniform, we transform the KL divergence minimization problem into entropy maximization problem. 
\subsection{Superiorities over VAE}

\paragraph{Why Uniform Prior?} Recently, \cite{gala} and \cite{gdn} have noticed that neural networks for structured data \eat{act as a low-pass filter in spectral domain,}usually cause spectral concentration, for instance, only low-frequency information is retained\eat{ and retains the commonality of node features}.\eat{ The smoothed features may work well for assortative networks.} However, \cite{lf1} verified high-frequency information \eat{which capture the heterogeneity between nodes }is also vital, especially for reconstruction networks.
Note that if we assume a Uniform distribution on the latent code representation,  the optimization objective tends to diversify the Laplacian distribution. Consequently, the phenomenon of spectral concentration is alleviated to some extent.

\paragraph{Why Deterministic Autoencoder?} Although MEGAE is equivalent to VAE from the perspective of optimization objective, there is still structural inconsistency between the two approaches as MEGAE is an autoencoder in nature. Why\eat{ don't we take the framework of variational autoencoders} not VAE? First,\eat{ we adopt the framework of autoencoder} MEGAE can avoid reparameterization in stochastic gradient descent. Assuming Uniform distribution as prior distribution, it is hard to perform reparameterization trick \cite{vae} for training\eat{, which is commonly adopted on Gaussian distribution in VAE}.
Second, the problem of posteriori collapse arises in VAE due to KL divergence vanishing\eat{posterior distribution closely matches the prior for a subset of latent variables}. \eat{Most existing works demonstrate that posterior collapse is caused by the KL-divergence term in the ELBO objective, leading the generative model to ignore a subset of the latent variables. }Since we take the maximum entropy principle as the basis, MEGAE successfully avoids this problem.

Finally, the certainty for each instance in autoencoder is beneficial to distinct imputation. Recently, \cite{vaeentropy} claims that deterministic encoders have been posited to retain high-frequency information, since they ‘lock’ codes into a single choice for each instance. The deterministic encoder can retain spectral information better than VAE.

\section{Experimental Setup}\label{app:exp}
\subsection{Dataset Summary}\label{app:data}
\textbf{As for multi-graph datasets}, we use dataset benchmarks from TUDatasets~\cite{tud} with continuous features of which the realization sizes vary from $N=10$ to 3848, and feature dimensions vary from $D=1$ to 780. The detailed information of datasets is summaried in Table~\ref{table6}.
\textbf{As for single-graph datasets}, we use four commonly used datasets. The detailed information of datasets is summaried in Table~\ref{table7}.
\begin{table}[ht]
  \renewcommand\arraystretch{1.2}

  \begin{center}

  \resizebox{\linewidth}{!}{
  \begin{tabular}{c|ccccc}
    \toprule
    Dataset & \thead{Mean\\ Nodes} & \thead{Mean\\Edges} & \thead{Features} & \thead{Graph\\ Number}  & \thead{Continuous\\data?} \\ 
    \midrule
    ENZYMES & 33 & 62 & 18 & 600  & $\surd$ \\
    PROTEINS\_full & 39 & 73 & 29 & 1,113  & $\surd$ \\
    QM9 & 18 & 19 & 16 & 1,290  & $\surd$ \\
    Synthie & 95 & 173 & 15 & 400  & $\surd$ \\
    FRANKENSTEIN & 17 & 18 & 780 & 4,337   & $\surd$ \\
    FIRSTMM\_DB & 1,377 & 3,074 & 1 & 41   & $\surd$ \\
  \bottomrule
  \end{tabular}
  }
  \end{center}
   \caption{Summary of multi-graph datasets.} 
     \label{table6}
\end{table}
\begin{table}[ht]
  \renewcommand\arraystretch{1.2}

  \begin{center}

  \resizebox{\linewidth}{!}{
  \begin{tabular}{c|ccccc}
    \toprule
    Dataset & \thead{Type} & \thead{Edges} & \thead{Nodes} & \thead{Classes} & \thead{Features} \\ 
    \midrule
    Cora & Citation & 5,069 & 2,485 & 7 & 1,433 \\
    Citeseer & Citation & 3,679 & 2,120 & 6 & 3,703  \\
    Pubmed & Citation & 44,324 & 19,717 & 3 & 500  \\
    AmaPhoto & Co-purchase & 119,043 & 7,487 & 8 & 745  \\
   
  \bottomrule
  \end{tabular}
  }

  \end{center}
   \caption{Summary of single-graph datasets.} 
   \label{table7} 
\end{table}
\subsection{Setup}\label{app:setup}
\paragraph{Multi-graph Datasets.}We use an 70-10-20 train-validation-test split and construct random missingness only on the test set. Each run has a different train-test split and the mask for feature missingness. Unless otherwise stated, missingness is applied by randomly removing features across all dimensions at a rate of 10\%. We run 5 trials with different seeds and report the Root Mean Squared Error (RMSE) results. We perform a hyperparameter sweep for $M$ and $\gamma$ as they change between $3$ and $20$ , $1e-2$ and $10$, respectively. After imputation, re-run the previous 5 trials to evaluate MEGAE for graph classification on the imputed test set. For all baselines, we use a 2-layer GCN~\cite{gcn} as a downstream GNNs. Finally, we report the results of mean RMSE and mean accuracy on the test set.
\paragraph{Single-graph Datasets.}We closely follow \cite{gcn} to perform standard dataset split. As for each class, 20 nodes are randomly chosen for training. Moreover, we choose 500 labeled nodes for validation and evaluate the prediction accuracy on a test set of 1,000 labeled nodes. For AmaPhoto, we follow the work~\cite{gcnmf} to randomly choose 40 nodes for each class for training, 500 nodes for validation and the remaining nodes for testing. For all datasets, each run has a different train-validation-test split and the mask for random missingness across all feature dimensions. A 2-layer GCN~\cite{gcn} is used as a downstream GNNs for node classification. After running for 5 trails, we report the results of mean RMSE for imputation and mean accuracy on the test set for node classification.
\subsection{Baseline Implementations}\label{app:baseline}
For baselines including GAIN, OT, MIRACLE, GraphVAE, MolGAN, GRAPE and GDN, we used the official implementation released by the authors on Github with all the same hyper-parameters as those in the source code. For MEAN and SVD, we used the implementation provided in the \textit{fancyimpute} package. For KNN, we used the implementation provided in the \textit{sklearn} module.
\begin{itemize}[leftmargin=*]
    \item Statistical imputation methods
    \begin{itemize}
         \item \textbf{MEAN}: The method imputes the missing value with the mean of feature variable. \url{https://github.com/iskandr/fancyimpute}
        \item \textbf{KNN}~\cite{knn}: The method imputes the missing value with weighted average of its K-nearest neighbors that can be observed. \url{https://scikit-learn.org/stable/modules/impute.html}
        \item \textbf{SVD}~\cite{svd}: The method replaces missing values with iterative low-rank SVD decomposition. \url{https://github.com/iskandr/fancyimpute}
    \end{itemize}
    
    \item Learning-based imputation methods
    \begin{itemize}
        \item \textbf{MICE}~\cite{mice}: A multiple regression method with Markov chain Monte Carlo theory. \url{https://github.com/amices/mice}
        \item \textbf{GAIN}~\cite{gain}: An imputation method with a generative adversarial network. \url{https://github.com/jsyoon0823/GAIN}
          \item \textbf{OT}~\cite{ot}: An end-to-end learning method using optimal transport. \url{https://github.com/BorisMuzellec/MissingDataOT}
        \item \textbf{MIRACLE}~\cite{miracle}: A state-of-the-art imputation method based on causal regularization. We report the best imputation result with refinements of baselines \textbf{MICE}, \textbf{GAIN} and \textbf{OT}. \url{https://github.com/vanderschaarlab/MIRACLE}
    \end{itemize}
    
    \item Graph learning methods
    \begin{itemize}
        \item \textbf{GraphVAE}~\cite{graphvae}: A method for reconstructing graphs with a generative model. \url{https://github.com/JiaxuanYou/graph-generation/tree/master/baselines/graphvae}
        \item \textbf{MolGAN}~\cite{molgan}: A state-of-the-art method for graph reconstruction. We report the imputation results based on the reconstructed graph features. \url{https://github.com/nicola-decao/MolGAN}
        \item \textbf{GRAPE}~\cite{grape}: A state-of-the-art imputation model with bipartite graph representation. \url{http://snap.stanford.edu/grape}
        \item \textbf{GDN}~\cite{gdn}: A method for recovering graph features from smoothed representations. \url{https://proceedings.neurips.cc/paper/2021/hash/afa299a4d1d8c52e75dd8a24c3ce534f-Abstract.html}
        \item \textbf{VGAE}~\cite{vgae}: A variational-based graph autoencoder for graph generation and reconstruction.
        \url{https://github.com/DaehanKim/vgae_pytorch}
        \item \textbf{GCNMF}~\cite{gcnmf}: A learnable Gaussian model for node classification with incomplete graph attributes.
        \url{https://github.com/marblet/GCNmf}
        \item \textbf{PaGNN}~\cite{pagnn}: A partial GNN which applies a partial message-passing scheme for incomplete graph data learning without feature imputation.
        \url{https://github.com/zyang1580/PAGNN_}
        \item \textbf{sRMGCNN}~\cite{mgcnn}: A geometric graph learning method for reconstructing missing attributes with observed features and graph structures.
        \url{https://github.com/fmonti/mgcnn}
        \item \textbf{GC-MC}~\cite{gcmc}: An imputation method with differentiable message passing on the bipartite interaction graphs.
        \url{https://github.com/tanimutomo/gcmc}
    \end{itemize}
\end{itemize}
\subsection{Computational Hardware.}
All models are trained with the following settings:
\begin{itemize}
    \item Operating system: Linux Red Hat 4.8.2-16
    \item CPU: Inter(R) Xeon(R) Platinum 8255C CPU @ 2.50GHz
    \item GPU: NVIDIA Tesla V100 (16G)
\end{itemize}
\section{Experiments under three missing mechanisms}\label{app:mar}
The missing data mechanisms can be categorized into three types~\cite{rubin1976inference}: (1) Missing Completely At Random (MCAR) - the probability of being missing is the same across all features; (2) Missing At Random (MAR) - the probability of being missing is the same within groups determined by observed values; (3) Missing Not At Random (MNAR) - the probability of being missing varies for unknown reasons. In this work, we aim to propose the method which is mainly applicable to MCAR and verify MEGAE's generalizability on the other two mechanisms with 30\% missing features.

\begin{figure*}[ht]
\centering
  \includegraphics[width=0.85\linewidth]{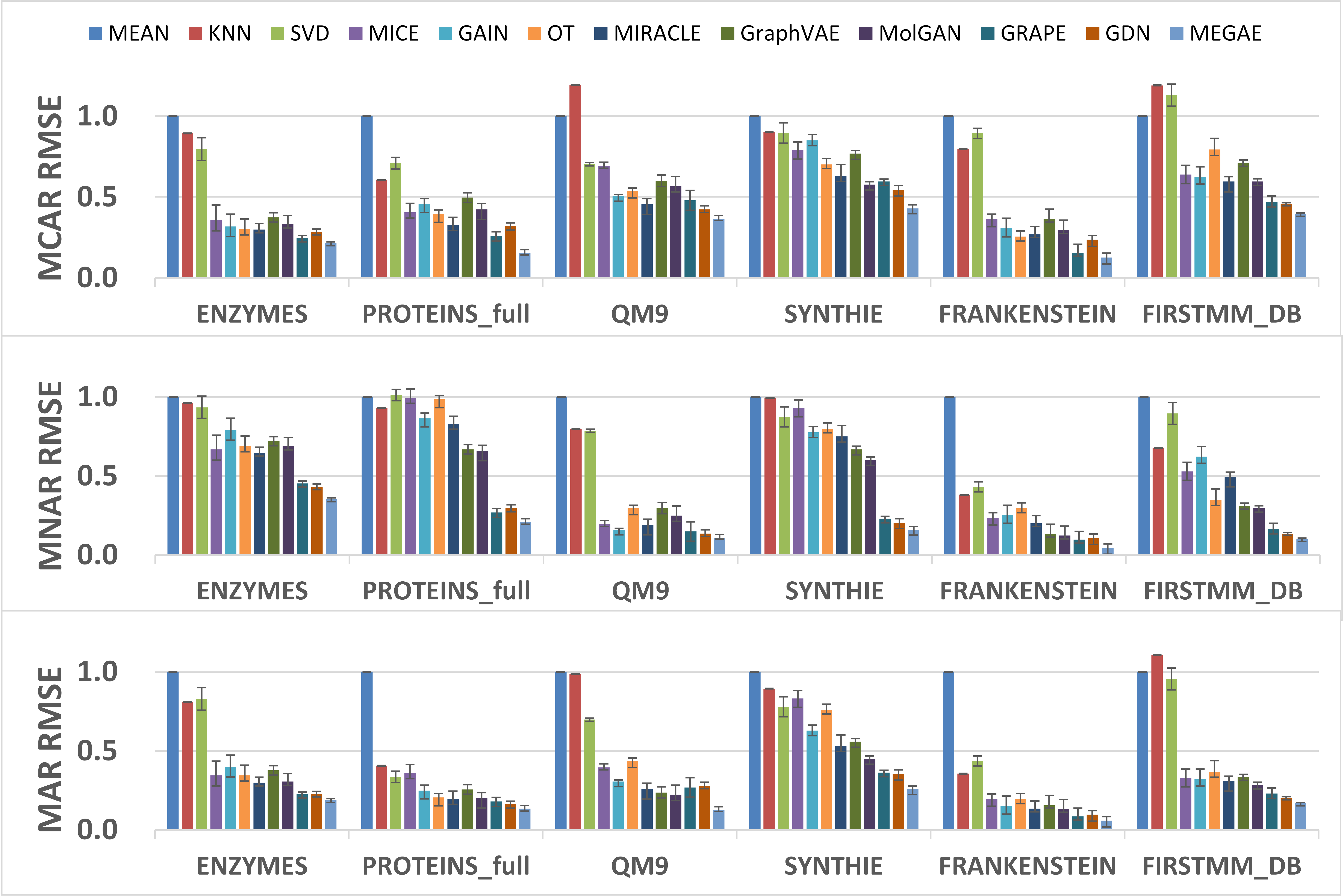}

\caption{
Results on 6 datasets at 30\% \textbf{MCAR} (\textit{top}), \textbf{MNAR} (\textit{middle}), and \textbf{MAR} (\textit{bottom}). All results are normalized by the \textbf{MEAN} imputation RMSE in each dataset. 
}
\label{figure3}
\end{figure*}

Using our experimental setup for multi-graph datasets in the main paper, we show the performance of MEGAE in terms of RMSE in Figure~\ref{figure3}.
\section{RMSE and accuracy results on Cora and Citeseer.}\label{app:cora}
Using our experimental setup for single-graph datasets in the main paper, we show the performance of MEGAE in terms of RMSE~(see Table~\ref{table8}) and accuracy~(see Table~\ref{table9}) on Cora and Citeseer.
\begin{table*}[t]

  \renewcommand\arraystretch{0.9}
\setlength\tabcolsep{2pt}
\centering
\begin{tabular}{@{}c|ccccc|ccccc@{}}
\toprule
\multirow{2}{*}{Methods} & \multicolumn{5}{c|}{Cora (RMSE)} & \multicolumn{5}{c}{CiteSeer (RMSE)} \\ \cmidrule(l){2-11} 
 & \multicolumn{1}{c|}{0.1 miss} & \multicolumn{1}{c|}{0.3 miss} & \multicolumn{1}{c|}{0.5 miss} & \multicolumn{1}{c|}{0.7 miss} & 0.99 miss & \multicolumn{1}{c|}{0.1 miss} & \multicolumn{1}{c|}{0.3 miss} & \multicolumn{1}{c|}{0.5 miss} & \multicolumn{1}{c|}{0.7 miss} & 0.99 miss \\ \midrule
GCNMF & \multicolumn{1}{c|}{-} & \multicolumn{1}{c|}{-} & \multicolumn{1}{c|}{-} & \multicolumn{1}{c|}{-} & - & \multicolumn{1}{c|}{-} & \multicolumn{1}{c|}{-} & \multicolumn{1}{c|}{-} & \multicolumn{1}{c|}{-} & - \\
PaGNN & \multicolumn{1}{c|}{-} & \multicolumn{1}{c|}{-} & \multicolumn{1}{c|}{-} & \multicolumn{1}{c|}{-} & - & \multicolumn{1}{c|}{-} & \multicolumn{1}{c|}{-} & \multicolumn{1}{c|}{-} & \multicolumn{1}{c|}{-} & - \\
sRMGCNN & \multicolumn{1}{c|}{0.1180} & \multicolumn{1}{c|}{0.1187} & \multicolumn{1}{c|}{0.1193} & \multicolumn{1}{c|}{0.1643} & \underline{0.1837} & \multicolumn{1}{c|}{0.0693} & \multicolumn{1}{c|}{\underline{0.0745}} & \multicolumn{1}{c|}{0.1137} & \multicolumn{1}{c|}{\underline{0.1163}} & \underline{0.1750} \\
GC-MC & \multicolumn{1}{c|}{0.0995} & \multicolumn{1}{c|}{0.1089} & \multicolumn{1}{c|}{0.1292} & \multicolumn{1}{c|}{0.1571} & 0.2352 & \multicolumn{1}{c|}{\underline{0.0599}} & \multicolumn{1}{c|}{0.0865} & \multicolumn{1}{c|}{0.1032} & \multicolumn{1}{c|}{0.1248} & 0.1892 \\
GRAPE & \multicolumn{1}{c|}{0.0975} & \multicolumn{1}{c|}{0.1049} & \multicolumn{1}{c|}{0.1256} & \multicolumn{1}{c|}{0.1359} & 0.2274 & \multicolumn{1}{c|}{0.0657} & \multicolumn{1}{c|}{0.0930} & \multicolumn{1}{c|}{0.1068} & \multicolumn{1}{c|}{0.1295} & 0.1926 \\
VGAE & \multicolumn{1}{c|}{0.1105} & \multicolumn{1}{c|}{0.1139} & \multicolumn{1}{c|}{0.1616} & \multicolumn{1}{c|}{0.2095} & 0.2892 & \multicolumn{1}{c|}{0.0774} & \multicolumn{1}{c|}{0.1060} & \multicolumn{1}{c|}{0.1056} & \multicolumn{1}{c|}{0.1350} & 0.2172 \\
GDN & \multicolumn{1}{c|}{\underline{0.0946}} & \multicolumn{1}{c|}{\underline{0.0964}} & \multicolumn{1}{c|}{\underline{0.1085}} & \multicolumn{1}{c|}{\underline{0.1332}} & 0.2037 & \multicolumn{1}{c|}{\underline{0.0599}} & \multicolumn{1}{c|}{0.0895} & \multicolumn{1}{c|}{\underline{0.0893}} & \multicolumn{1}{c|}{0.1240} & 0.1784 \\
\textbf{MEGAE} & \multicolumn{1}{c|}{\textbf{0.0804}} & \multicolumn{1}{c|}{\textbf{0.0849}} & \multicolumn{1}{c|}{\textbf{0.0878}} & \multicolumn{1}{c|}{\textbf{0.0941}} & \textbf{0.1208} & \multicolumn{1}{c|}{\textbf{0.0567}} & \multicolumn{1}{c|}{\textbf{0.0621}} & \multicolumn{1}{c|}{\textbf{0.0741}} & \multicolumn{1}{c|}{\textbf{0.0938}} & \textbf{0.1408} \\ \midrule
\multirow{3}{*}{\begin{tabular}[c]{@{}c@{}}Performance\\       gain\end{tabular}} & \multirow{3}{*}{\begin{tabular}[c]{@{}c@{}}0.0142\\      |\\      0.0376\end{tabular}} & \multirow{3}{*}{\begin{tabular}[c]{@{}c@{}}0.0115\\      |\\      0.0338\end{tabular}} & \multirow{3}{*}{\begin{tabular}[c]{@{}c@{}}0.0207\\      |\\      0.0738\end{tabular}} & \multirow{3}{*}{\begin{tabular}[c]{@{}c@{}}0.0392\\      |\\      0.1154\end{tabular}} & \multirow{3}{*}{\begin{tabular}[c]{@{}c@{}}0.0629\\      |\\      0.1884\end{tabular}} & \multirow{3}{*}{\begin{tabular}[c]{@{}c@{}}0.0032\\      |\\      0.0207\end{tabular}} & \multirow{3}{*}{\begin{tabular}[c]{@{}c@{}}0.0124\\      |\\      0.0439\end{tabular}} & \multirow{3}{*}{\begin{tabular}[c]{@{}c@{}}0.0152\\      |\\      0.0396\end{tabular}} & \multirow{3}{*}{\begin{tabular}[c]{@{}c@{}}0.0225\\      |\\      0.0413\end{tabular}} & \multirow{3}{*}{\begin{tabular}[c]{@{}c@{}}0.0342\\      |\\      0.0764\end{tabular}} \\
 &  &  &  &  &  &  &  &  &  &  \\
 &  &  &  &  &  &  &  &  &  &  \\ \bottomrule
\end{tabular}%
\caption{Mean RMSE results of attribute imputation with different missing rates on Cora and Citeseer.}
 \label{table8}
\end{table*}
\begin{table*}[t]

  \renewcommand\arraystretch{0.9}
\setlength\tabcolsep{2pt}
\centering
\begin{tabular}{@{}c|ccccc|ccccc@{}}
\toprule
\multirow{2}{*}{Methods} & \multicolumn{5}{c|}{Cora (Acc. = 81.41\%   with full attributes)} & \multicolumn{5}{c}{CiteSeer (Acc. =   67.14\% with full attributes)} \\ \cmidrule(l){2-11} 
 & \multicolumn{1}{c|}{0.1 miss} & \multicolumn{1}{c|}{0.3 miss} & \multicolumn{1}{c|}{0.5 miss} & \multicolumn{1}{c|}{0.7 miss} & 0.99 miss & \multicolumn{1}{c|}{0.1 miss} & \multicolumn{1}{c|}{0.3 miss} & \multicolumn{1}{c|}{0.5 miss} & \multicolumn{1}{c|}{0.7 miss} & 0.99 miss \\ \midrule
GCNMF & \multicolumn{1}{c|}{\underline{80.72\%}} & \multicolumn{1}{c|}{79.87\%} & \multicolumn{1}{c|}{\underline{77.71\%}} & \multicolumn{1}{c|}{73.26\%} & 32.93\% & \multicolumn{1}{c|}{66.34\%} & \multicolumn{1}{c|}{65.65\%} & \multicolumn{1}{c|}{64.37\%} & \multicolumn{1}{c|}{60.09\%} & 29.77\% \\
PaGNN & \multicolumn{1}{c|}{80.51\%} & \multicolumn{1}{c|}{79.44\%} & \multicolumn{1}{c|}{77.47\%} & \multicolumn{1}{c|}{74.57\%} & 33.90\% & \multicolumn{1}{c|}{\underline{66.49\%}} & \multicolumn{1}{c|}{\underline{65.88\%}} & \multicolumn{1}{c|}{64.23\%} & \multicolumn{1}{c|}{60.85\%} & 27.48\% \\
sRMGCNN & \multicolumn{1}{c|}{78.93\%} & \multicolumn{1}{c|}{77.72\%} & \multicolumn{1}{c|}{76.62\%} & \multicolumn{1}{c|}{74.96\%} & 28.88\% & \multicolumn{1}{c|}{65.89\%} & \multicolumn{1}{c|}{65.22\%} & \multicolumn{1}{c|}{63.02\%} & \multicolumn{1}{c|}{\underline{61.27\%}} & 28.84\% \\
GC-MC & \multicolumn{1}{c|}{79.62\%} & \multicolumn{1}{c|}{79.03\%} & \multicolumn{1}{c|}{75.59\%} & \multicolumn{1}{c|}{73.32\%} & 20.03\% & \multicolumn{1}{c|}{66.12\%} & \multicolumn{1}{c|}{64.81\%} & \multicolumn{1}{c|}{63.92\%} & \multicolumn{1}{c|}{61.07\%} & 36.61\% \\
GRAPE & \multicolumn{1}{c|}{80.52\%} & \multicolumn{1}{c|}{79.53\%} & \multicolumn{1}{c|}{76.02\%} & \multicolumn{1}{c|}{74.52\%} & 27.62\% & \multicolumn{1}{c|}{66.32\%} & \multicolumn{1}{c|}{64.27\%} & \multicolumn{1}{c|}{63.28\%} & \multicolumn{1}{c|}{59.82\%} & \underline{38.42\%} \\
VGAE & \multicolumn{1}{c|}{78.82\%} & \multicolumn{1}{c|}{78.02\%} & \multicolumn{1}{c|}{73.24\%} & \multicolumn{1}{c|}{70.06\%} & 14.57\% & \multicolumn{1}{c|}{65.48\%} & \multicolumn{1}{c|}{63.95\%} & \multicolumn{1}{c|}{63.46\%} & \multicolumn{1}{c|}{56.78\%} & 16.49\% \\
GDN & \multicolumn{1}{c|}{80.70\%} & \multicolumn{1}{c|}{\underline{80.03\%}} & \multicolumn{1}{c|}{77.22\%} & \multicolumn{1}{c|}{\underline{75.67\%}} & \underline{34.54}\% & \multicolumn{1}{c|}{66.25\%} & \multicolumn{1}{c|}{\underline{64.37\%}} & \multicolumn{1}{c|}{63.76\%} & \multicolumn{1}{c|}{60.20\%} & 32.62\% \\
\textbf{MEGAE} & \multicolumn{1}{c|}{\textbf{80.99\%}} & \multicolumn{1}{c|}{\textbf{80.80\%}} & \multicolumn{1}{c|}{\textbf{79.01\%}} & \multicolumn{1}{c|}{\textbf{78.77\%}} & \textbf{42.44\%} & \multicolumn{1}{c|}{\textbf{66.94\%}} & \multicolumn{1}{c|}{\textbf{66.29\%}} & \multicolumn{1}{c|}{\textbf{64.98\%}} & \multicolumn{1}{c|}{\textbf{62.81\%}} & \textbf{45.39\%} \\ \midrule
\multirow{3}{*}{\begin{tabular}[c]{@{}c@{}}Performance\\       gain\end{tabular}} & \multirow{3}{*}{\begin{tabular}[c]{@{}c@{}}0.27\%\\      |\\      2.17\%\end{tabular}} & \multirow{3}{*}{\begin{tabular}[c]{@{}c@{}}0.77\%\\      |\\      3.08\%\end{tabular}} & \multirow{3}{*}{\begin{tabular}[c]{@{}c@{}}1.3\%\\      |\\      5.77\%\end{tabular}} & \multirow{3}{*}{\begin{tabular}[c]{@{}c@{}}3.10\%\\      |\\      8.71\%\end{tabular}} & \multirow{3}{*}{\begin{tabular}[c]{@{}c@{}}7.90\%\\      |\\      27.87\%\end{tabular}} & \multirow{3}{*}{\begin{tabular}[c]{@{}c@{}}0.45\%\\      |\\      1.46\%\end{tabular}} & \multirow{3}{*}{\begin{tabular}[c]{@{}c@{}}0.41\%\\      |\\      2.34\%\end{tabular}} & \multirow{3}{*}{\begin{tabular}[c]{@{}c@{}}0.61\%\\      |\\      1.96\%\end{tabular}} & \multirow{3}{*}{\begin{tabular}[c]{@{}c@{}}1.54\%\\      |\\      6.03\%\end{tabular}} & \multirow{3}{*}{\begin{tabular}[c]{@{}c@{}}6.97\%\\      |\\      28.90\%\end{tabular}} \\
 &  &  &  &  &  &  &  &  &  &  \\
 &  &  &  &  &  &  &  &  &  &  \\ \bottomrule
\end{tabular}%
\caption{Test accuarcy with different missing rates on Cora and Citeseer node classification benchmarks.}
 \label{table9}
\end{table*}

\end{document}